\documentclass[10pt,twocolumn,letterpaper]{article}

\usepackage{cvpr}              % To produce the CAMERA-READY version
\usepackage{graphicx}
\usepackage{amsmath}
\usepackage{amssymb}
\usepackage{booktabs}
\usepackage[table]{xcolor}% http://ctan.org/pkg/xcolor
\usepackage{multirow}

\usepackage{algorithm}
\usepackage{algorithmic}
\usepackage{subcaption}
\usepackage{mathtools}
\usepackage{enumitem}

\usepackage[pagebackref,breaklinks,colorlinks]{hyperref}

\usepackage[capitalize]{cleveref}
\crefname{section}{Sec.}{Secs.}
\Crefname{section}{Section}{Sections}
\Crefname{table}{Table}{Tables}
\crefname{table}{Tab.}{Tabs.}

\def\cvprPaperID{705} % *** Enter the CVPR Paper ID here
\def\confName{CVPR}
\def\confYear{2023}

\newcommand{\bI}{\mathbf{I}}

\newcommand{\bx}{\mathbf{x}}
\newcommand{\by}{\mathbf{y}}
\newcommand{\bz}{\mathbf{z}}

\begin{document}

\title{Bi-Noising Diffusion: Towards Conditional Diffusion Models \\ with Generative Restoration Priors}
\author{
Kangfu Mei\\
Johns Hopkins University\\
{\tt\small kmei1@jhu.edu}
\and
Nithin Gopalakrishnan Nair\\
Johns Hopkins University\\
{\tt\small ngopala2@jhu.edu}
\and
Vishal M. Patel\\
Johns Hopkins University\\
{\tt\small vpatel36@jhu.edu}
}
\maketitle

%%%%%%%%% ABSTRACT
\begin{abstract}
Conditional diffusion probabilistic models can model the distribution of natural images and can generate diverse and realistic samples based on given conditions. However, oftentimes their results can be unrealistic with observable color shifts and textures. We believe that this issue results from the divergence between the probabilistic distribution learned by the model and the distribution of natural images. The delicate conditions gradually enlarge the divergence during each sampling timestep. To address this issue, we introduce a new method that brings the predicted samples to the training data manifold using a pretrained unconditional diffusion model. The unconditional model acts as a regularizer and reduces the divergence introduced by the conditional model at each sampling step. We perform comprehensive experiments to demonstrate the effectiveness of our approach on super-resolution, colorization, turbulence removal, and image-deraining tasks. The improvements obtained by our method suggest that the priors can be incorporated as a general plugin for improving conditional diffusion models.
Our demo is \href{https://kfmei.page/bi-noising/}{https://kfmei.page/bi-noising/}.
\end{abstract}

%%%%%%%%% BODY TEXT

\section{Introduction}
In recent years, conditional image generation has received significant attention in the computer vision community. Some applications that make use of conditional image generation include text-to-image generation (\eg DALLE-2~\cite{ramesh2022hierarchical}) and image restoration (\eg SR3~\cite{saharia2021image}). The most challenging part of these restoration applications comes from the ill-posedness, \ie, the same degraded images may come from multiple different ground truth images.
The ill-posedness affects the performance of traditional methods like sparse coding~\cite{mairal2007sparse, mairal2009non} and makes it difficult for the learning-based algorithms to solve this problem.
Although recent learning-based methods have made impressive progress ~\cite{ledig2017photo}, there remains a significant quality gap between the prediction and natural images.

Recent works that utilize pretrained generative networks have shown the superior visual quality of conditional generation compared to the aforementioned end-to-end learning methods.
Generative models have shown impressive image generation results in terms of sample quality and diversity, indicating their capacity for encapsulating rich photorealistic priors.
Some representative methods include Generative Adversarial Networks (GANs)~\cite{goodfellow2014generative}, Variational Autoencoders (VAEs)~\cite{kingma2013auto}, and Autoregressive models~\cite{larochelle2011neural}.
Their generation process generally starts from the standard normal distribution from which diverse high-fidelity images sampled ~\cite{karras2019style, karras2020analyzing}.
Recent work~\cite{richardson2021encoding} has shown that the \emph{continuity} in the normal distribution remains preserved in the sampled results.
For example, the results produced from two different Gaussian noises with the same model will be close to each other if the two noises are close to each other in Euclidean space.
The continuity allows one to perform conditional image generation in an inversion manner that inverts degraded images into standard noises.
This inverted noise can then generate clear images by projecting the noise with generative models.
Following the protocol, multiple GAN-based generative priors, including optimization-based~\cite{menon2020pulse} and learning-based~\cite{richardson2021encoding} schemes have been proposed for various real-world tasks~\cite{wang2021towards}.

\begin{figure*}[ht!]
    \center
    \includegraphics[width=\linewidth]{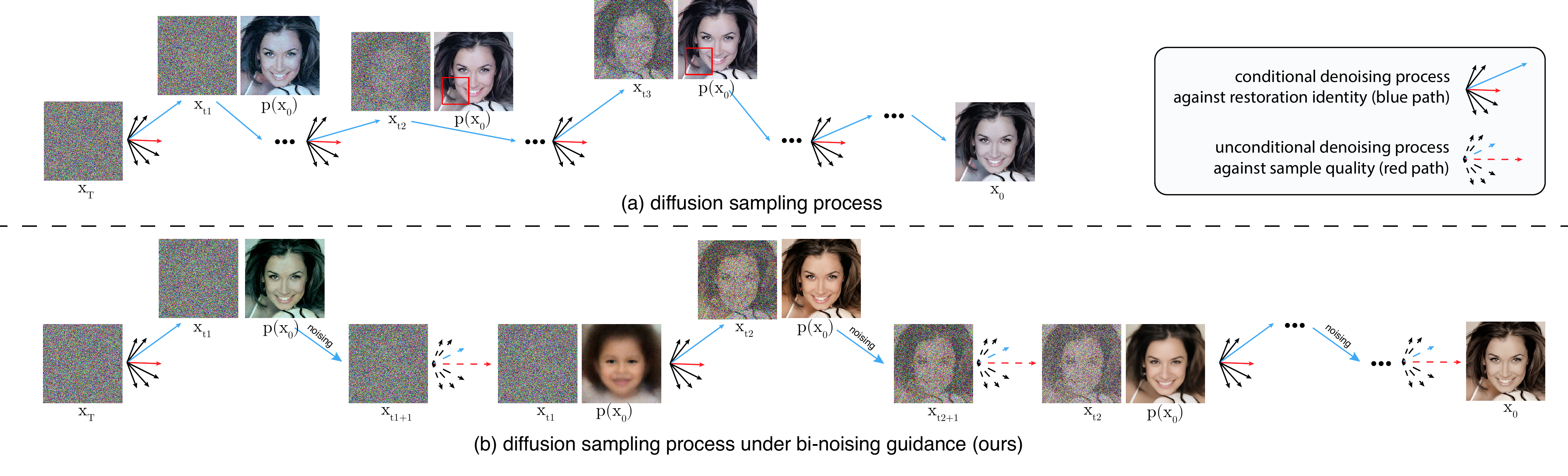}
    \vspace{-1.5\baselineskip}
    \caption{The graphical model showing the difference between the previous diffusion sampling process and ours with bi-noising guidance for colorization, where $\bx_t$ is the noise of each diffusion process at timestep $t$, $p(\bx_0)$ is the predicted noise-free start point of $\bx_0$, and arrows indicate the denoising results of the diffusion models at each denoising process.
    Top figure shows how the conditional denoising process for colorization gradually accumulates the incorrect noise and results in artifacts.  Instead, as shown on the bottom figure, the proposed additional noising and denoising steps diminish the incorrect noise and help in achieving better results.}
    \label{fig:teaser}
    \vspace{-.5\baselineskip}
\end{figure*}

Denoising diffusion probabilistic models~\cite{sohl2015deep, ho2020denoising} are the most recent deep generative models.
They have shown comparable and even better performance at image synthesis than GANs with delicate guidance~\cite{dhariwal2021diffusion}.
These models learn to sequentially denoise stochastic noise map starting from the normal distribution $\mathcal{N}(0, \mathbf{I})$ to clean images.
However, the generation process is stochastic, and the continuity cannot be preserved from the initial sampled noise.
For instance, two sampled noises from the same normal distribution with a small divergence may generate significantly different clear images.
Such a noncontinuous generation process prevents the generative priors from being applied along with the denoising process like the inversion GANs~\cite{richardson2021encoding}. 
Hence, despite their impressive synthesis capacity, diffusion model-based priors have not been explored before.

In this work, we introduce a new method, named bi-noising diffusion, for utilizing rich priors encapsulated in the unconditional pretrained diffusion models.
Inspired by implicit sampling that was first developed in the denoising diffusion implicit models~\cite{song2020denoising} for acceleration, we show that the implicit sampling using an unconditional pretrained diffusion model has a capacity for correcting the divergence of distributions modeled by the conditional diffusion models. Specifically, we make a coarse implicit prediction at each intermediate diffusion time step by sampling from the conditional model.
We then sample the prediction back to the intermediate step with the forward diffusion process. Finally, we make a refined prediction by utilizing an unconditional model.
Fig.~\ref{fig:teaser} visualizes the bi-noising procedure and the error by predicting the noiseless start-point image $p(\bx_0)$ of the noise image $\bx_t$.
Using this two-step procedure, one can utilize the embedded rich priors learned by the unconditional model and produce better-quality images. This hypothesis is further validated through extensive experiments demonstrating that the introduced method performs favorably against state-of-the-art conditional diffusion models.

\section{Related Work}
% Diffusion models belong to the family of variational models~\cite{van2016pixel, germain2015made, lecun2006tutorial, kingma2013auto, rezende2014stochastic}, which learn the distribution density directly by maximizing the likelihood of data distribution.
% In contrast, another widely used generative model, i.e., Generative Adversarial Networks~\cite{goodfellow2014generative} (GANs) belong to the family of implicit generative models, which implicitly represent the data distribution in its sampling process.
% Motivated by the impressive generative capability of GANs, the research community has proposed various works on extracting priors from pre-trained GANs for conditional image generation, while methods based on diffusion models focus on classifier guidance fashion.
% In what follows, we discuss the representative methods for extracting priors based on GANs. For a detailed review of these methods, the reader is referred to Xia et al.~\cite{xia2021gan}.

\noindent\textbf{Iterative methods.} Finding the corresponding latent code~\cite{creswell2018inverting, abdal2019image2stylegan, abdal2020image2stylegan++, menon2020pulse, gu2020image} or sampled noise~\cite{choi2021ilvr, meng2022sdedit, lugmayr2022repaint} of distorted images for restoration is one of the most straightforward ways of utilizing the generative priors.
The intuition is that the pretrained generative models tend to produce natural results from their initial distribution. Thus the corresponding latent code or sampled noise can be projected to the restored images without additional optimization or learning.
Menon et al.~\cite{menon2020pulse} proposed optimizing the latent code based on the difference between the generative results and the distorted images.
Gu et al.~\cite{gu2020image} proposed to optimize multiple latent codes and compose them together for better visual quality.
Similar iterative methods based on diffusion models have also been explored.
Choi et al.~\cite{choi2021ilvr} proposed to refine the sampled noise at each reverse diffusion step with the residual of distorted images.
However, the applied stochastic iterative process tends to produce significantly different results though slightly changing its input.
Therefore, these methods can only be applied to applications that do not require preserving the image identity.

\noindent\textbf{Learning-based methods.} Employing additional encoders~\cite{zhu2016generative, richardson2021encoding, wang2021towards, chan2021glean} to predict the latent code is another promising way that can bypass the stochastic optimization issues.
However, such a method is incompatible with the diffusion models since it is impossible to encode the distribution of each reverse diffusion process for models that employ many sampling timesteps.
Existing works learn to model conditional generative restoration~\cite{saharia2021image, whang2021deblurring} instead.
Richardson et al.~\cite{richardson2021encoding} proposed to encode images with a ResNet backbone into an extended $\mathcal{W}+$ latent space, which defines upon features of each input layer of the generative networks.
Wang et al.~\cite{wang2021towards} proposed to encode images with a U-Net backbone and modulate the features of each generative layer of the generative networks.
Saharia et al.~\cite{saharia2021image} proposed to learn the noise distribution with the distorted image as the condition.
Whang et al.~\cite{whang2021deblurring} proposed to learn the generative process of residual given restored images.
Compared with these GAN-based learning methods, a large number of sampling timesteps significantly increases the complexity of designing the corresponding encoders and thus makes the priors difficult to be learned.

\noindent\textbf{Classifier guidance.}
Diffusion models have been using class information heavily to perform truncated or low-temperature sampling to increase the sample quality.
The initial attempt~\cite{sohl2015deep, song2020score, dhariwal2021diffusion, nichol2021glide} is to incorporate a pre-trained classifier by using its gradients to guide the diffusion sampling process.
However, it complicates the diffusion model because additional training is required for the classifier on noisy data.
Classifier-free guidance~\cite{ho2022classifier, wang2022semantic} is another approach for addressing the complexity issue.
It alleviates the complexity by combining the existing network with the classifier for guidance, \eg, Ho et al.~\cite{ho2022classifier} use conditional diffusion network with an empty condition, and Wang et al.~\cite{wang2022semantic} use pretrained segmentation with a null label.
% In a nutshell, the guided diffusion process is
% $(1+w)\epsilon_\theta(\bx, c) - w\epsilon_\theta(\bx, \varnothing)$ with a strength control $w$ .
Nevertheless, the classifier fails at natural images, and its gradient is meaningless for restoration.
Its strength parameters also become less reasonable for the almost definite restoration sampling process.
In this paper, we are interested in incorporating the sampling quality superiority of the empty condition and the sampling guidance ability of degraded images.
We show that the empty condition can bring the incorrect noisy image back into the high-quality manifold.
Compared with the classifier and classifier-free guidance,
our bi-noising guided diffusion process keeps the same complexity but better fits the restoration task.

\section{Proposed Method}
In this section, we discuss the proposed mechanism to add the embedded priors to diffusion models. 
For consistency, we denote the intermediate output of the unconditional diffusion model as $\boldsymbol{\epsilon_\theta}(\cdot)$, parameterized by $\theta$ in the upcoming discussions following Denoising Diffusion Probabilistic Models~\cite{ho2020denoising} (DDPM). The additional, conditional diffusion model is denoted by  $\boldsymbol{f_\phi}(\cdot)$, the condition (\ie degraded images) and natural image pairs are denoted by $\{\bx_0, \by_0\}$, where the conditional diffusion model $\boldsymbol{f_\phi}(\cdot)$ with parameters $\phi$ denoises noisy image $\bx_t$ at timestep $t$ with the concatenated condition $\by_0$.

\subsection{Preliminaries}
Diffusion probabilistic models belong to a new family of generative models~\cite{sohl2015deep, song2019generative, ho2020denoising, nichol2021improved, dhariwal2021diffusion} that can effectively model intractable distributions~\cite{sohl2015deep}. A diffusion process consists of two parts, \ie, the forward process and the reverse diffusion process. In the forward diffusion process, a clean image is sampled from its data distribution and destroyed in $T$ timesteps by repetitive noising using Gaussians of very small variances. Specifically, the forward process can be formulated as
\setlength{\belowdisplayskip}{0pt} \setlength{\belowdisplayshortskip}{0pt}
\setlength{\abovedisplayskip}{0pt} \setlength{\abovedisplayshortskip}{0pt}
\begin{align}
\begin{split}    
  q(\by_t | \by_{t-1}) &= \mathcal{N}\left(\by_t ; \sqrt{{\beta}_{t}} \by_0,\left(1-{\beta}_{t}\right) \mathbf{I}\right) \\
  &= \sqrt{{\beta}_{t}} \by_0 + \epsilon \sqrt{1-{\beta}_{t}}, \epsilon \sim \mathcal{N}(0, \mathbf{I}),
\end{split}
\end{align}
or
\begin{align}
\begin{split}  
  q(\by_t | \by_{0}) &= \mathcal{N}\left(\by_t ; \sqrt{\bar{\alpha}_{t}} \by_0,\left(1-\bar{\alpha}_{t}\right) \mathbf{I}\right) \\
  &= \sqrt{\bar{\alpha}_{t}} \by_0 + \epsilon \sqrt{1-\bar{\alpha}_{t}}, \epsilon \sim \mathcal{N}(0, \mathbf{I}),
\end{split}
\end{align}
where $\alpha_t =1 - \beta_t$ and $\bar \alpha_t := \prod^t_{s=1} \alpha_s$ come from the variance schedule $\{\beta_1, \dots, \beta_T\}$ . The key idea here is that for large values of $T$, repetitive noising using Gaussians of small variances lead to a standard Gaussian, \ie, 
\begin{align}
 q(\by_T | \by_{0}) &= \mathcal{N}\left(\by_T ;  0, \mathbf{I}\right). 
\end{align}

Now at each reverse timestep $t$, we attempt to reconstruct the noisy $y_{t-1}$ from $y_t$ using a distribution $p$ modeled by a neural network with parameters $\theta$. The parameters of the distribution $p_{\theta}(\cdot)$, found by optimizing variational lower bound of log-likelihood of  $p_{\theta}(y_0)$, which is simplified by Ho et al.~\cite{ho2020denoising} by claiming that the major component in  the objective comes from $L_{t-1}$, and the simplified loss is 
\begin{align}
L_{t-1} &= E_{t \sim[1, T], \epsilon \sim \mathcal{N}(0, \mathbf{I})}\left[\left\|\epsilon-\epsilon_{\theta}\left(\mathbf{y}_t, t\right)\right\|^{2}\right].
\end{align}
Here network $\epsilon_{\theta}(\cdot)$ models the noise $\epsilon \sim \mathcal{N}(0, \mathbf{I})$ at each timestep $t$ with the denoised one $\by_t$, which can be seen as the process of learning the gradient of distributions with score matching according to Song et al.~\cite{song2019generative}. 
Therefore, we can learn the impressive perceptual synthesizing capacity with the simplified loss function between noises.

% \begin{figure}[htbp]
% 	\centering
% 		\includegraphics[width=0.45\textwidth]{diffusion_prior.pdf}
% 		%\includegraphics[width=0.5\linewidth]{Network_architecture/transformer1.png}
% 	\centering
% \vskip -10pt	\caption{Illustration of the manifold correction step at an intermediate timestep. At any intermediate step, the output of the conditional model is corrected using an unconditional model to map back to the natural image manifold. Here noisy images are not used for clearer illustration.}
% 	\label{fig:manifold}
% \vspace{-4mm}
% \end{figure}

\subsection{Learning to Refine Diffusion Process}
In our experiments, we denote the recent diffusion models~\cite{saharia2021image, whang2021deblurring} that learn the diffusion process with conditions as the way of Learning to Refine Diffusion Process (LRDP).
LRDP models the conditional distribution of a clean image given a degraded image for restoration learning, and thus it requires separate training for different tasks or datasets.
The objective for this learning process is formulated as
\begin{align}
    L_{\text {vlb}}:=E_{t \sim[1, T], \epsilon \sim \mathcal{N}(0, \mathbf{I})}\left[\left\|\epsilon-\epsilon_{\theta}\left(\mathbf{x}_0, \mathbf{y}_t, t\right)\right\|^{2}\right],
\end{align}
where $\by_t \sim \mathcal{N}(\by_t | \sqrt{\bar \alpha_t} \by_0, (1-\bar \alpha_t)\mathbf{I})$.
The network architecture in LRDP is a slightly changed version from the original U-Net found in DDPM, and the additional input $\bx_0$ and $\by_0$ are concatenated and passed to the input layer.
Similarly, the reverse diffusion process of LRDP is slightly changed from the original one and formulated as
\begin{align}
    \by_{t-1}=\frac{1}{\sqrt{\alpha_{t}}}\left(\by_{t}-\frac{1-\alpha_{t}}{\sqrt{1-\bar{\alpha}_{t}}} \boldsymbol{\epsilon}_{\theta}\left(\bx_0, \by_{t}, t\right)\right)+\sqrt{1-\alpha_t} \bz,
\end{align}
where $\bz \sim \mathcal{N}(0, \bI)$ and $\alpha_t, \bar \alpha_t$ is the variant of the pre-defined variance schedule $\{\beta_1, \dots, \beta_T\}$, that is $\alpha_t := 1 - \beta_t$ and $\bar \alpha_t := \prod^t_{s=1} \alpha_s$.
Since the diffusion process conditions on the specific type of degradation $p(\cdot)$ that produces degraded image $\bx_0$ given clear image $\by_0$ as $p(\bx_0|\by_0)$, LRDP needs re-training from scratch for different restoration tasks, which further heightens the training cost.

We experimentally find that such a protocol degrades the visual quality of the generation compared with the one without $\bx_0$ conditioned.
The most straightforward assumption towards the performance drop is that the assumed posterior $p(\bx_0|\by_0)$ contrasts with the diffusion process $p_\theta(\by_0 | \bx_0) \propto p_\theta(\by_0) p_\theta(\bx_0|\by_0)$ due to the ambiguous property of the degradation models.
Thus, we claim that decomposing the diffusion generation process into different protocols should be a more promising way to handle restoration tasks.

\subsection{Conditioning on Diffusion Process}
The recent work~\cite{choi2021ilvr, lugmayr2022repaint} falls into another category of utilizing the diffusion process, which uses a pretrained DDPM and changes its reverse diffusion process with distorted images by Conditioning on Diffusion Process (CDP).
A similar way was previously explored in the other generative models, \eg, mGANprior~\cite{gu2020image} and PULSE~\cite{menon2020pulse} invert a trained GAN by optimizing its latent code.
However, CDP does not require optimization compared with the previously mentioned GAN-based methods.
In contrast, it ensembles the conditions during sampling as
\begin{align}
  \hat \by_{t-1} &=\frac{1}{\sqrt{\alpha_{t}}}\left(\by_{t}-\frac{1-\alpha_{t}}{\sqrt{1-\bar{\alpha}_{t}}} \boldsymbol{\epsilon}_{\theta}\left(\by_{t}, t\right)\right)+\sqrt{1-\alpha_t} \bz \\
  \by_{t-1} &= \hat \by_{t-1} + \boldsymbol{\sigma}(x_0, \hat \by_{t-1}),
\end{align}
where $\boldsymbol{\sigma}(\cdot)$ is a handcrafted transformation which aims at combining $\bx_0$ with $\hat \by_{t-1}$ for accurate restoration.
For example, Choi et al.~\cite{choi2021ilvr} proposed to downsample $\bx_0$ and $\hat \by_{t-1}$ and take their residual as the conditioning, while Lugmayr et al.~\cite{lugmayr2022repaint} proposed to sum the visible region of $\bx_0$ with the invisible region $\hat \by_{t-1}$ for the inpainting task.

Though CDP avoids the heavy training cost and is suitable for some conditional generation tasks like restoration with minimal modifications, its performance highly depends on the amount of degradation in the conditioned images.
For example, when the conditioned images suffer from high amounts of distortion for face image restoration, CDP cannot preserve the face identity and tends to generate pseudo-sharp results with fake details.
These fake details introduce further ill-posedness to the restored images and greatly limit the applications of such methods.
Therefore, we propose refining the denoised results for correcting such artifacts at each step.

\subsection{Implicit Error-feedback Diffusion Priors}
\label{section:forward}
Since the diffusion models follow a time-sequential process, the error in each step and the visual artifacts propagate and add up, hence severely degrading the quality of some CDP results. However, such issues are rarely observed in the unconditional diffusion models.
We argue that the difference comes from conditioning breaking the inherent probabilistic distribution of noises at each sampling timestep, causing them to deviate from the manifold of natural images. Therefore, we propose to apply generative priors embedded in a pretrained unconditional model to regularize the noise predicted at each timestep from the conditional model.
The trained diffusion model with conditioning denoted as $\boldsymbol{f_\phi}(\cdot)$ takes as input the predicted image of the previous timestep and makes an implicit prediction $\tilde{\by}_0$ defined by
\begin{align}
  \tilde\by_0 = (\by_t - \sqrt{1 - \bar \alpha_t} \boldsymbol{f_\phi}(\bx_0, \by_t, t)) / \sqrt{\bar \alpha_t}.
\end{align}

Here $y_t$ denotes the prediction at the previous timestep. We then estimate the noisy version of the implicit prediction, which undergoes further regularization from an unconditional diffusion model. Please note that the unconditional diffusion model that fits the inherent probabilistic distribution.
The diffusion process $\by_t \sim q(\by_t | \tilde{\by}_0)$ with $\boldsymbol{\epsilon_\theta}(\cdot)$ is formulated as
\begin{align}
\begin{split}
    q(\by_t | \tilde{\by}_0) &:= \mathcal{N}(\by_t | \sqrt{\bar \alpha_t} \tilde \by_0, (1-\bar \alpha_t) \mathbf{I})\\
  &= \sqrt{\bar{\alpha}_t} \tilde{\mathbf{y}}_0 + \epsilon \sqrt{1-\bar{\alpha}_{t}}, \epsilon \sim \mathcal{N}(0, \mathbf{I}),
\end{split}
\end{align}
and
\begin{align}
  \by_{t-1} = \frac{1}{\sqrt{\alpha_t}}\left(\by_t-\frac{1-\alpha_t}{\sqrt{1-\bar{\alpha}_t}} \boldsymbol{\epsilon_\theta}\left(\by_{t}, t\right)\right)+\sigma_{t} \mathbf{z}.
  \label{eq:predict}
\end{align}
Following this procedure brings in an inherent regularization to the output of the conditional model during the reverse diffusion process.
Note that Equation~\eqref{eq:predict} takes the noised version $\by_t$ sampled from $\tilde \by_0$ as input. It is similar to the original reverse diffusion process, which takes the noised version of natural images as input.

In summary, we utilize two diffusion models for conditional image generation. The unconditional diffusion model regularizes the predicted outliers at each prediction timestep of the conditional diffusion model in an error-feedback way.
Moreover, for the complex real-world application like draining where domain gaps may exist, we further discuss the details of applying our bi-noising diffusion with slight modifications to achieve better performance.

% This section will be added back

\subsection{Complex Application: Deraining}
Here we discuss one of the applications where we apply our introduced bi-noising diffusion for further clarification.
% Several utilities are introduced for achieving state-of-the-art performance on the complex real-world task with the priors.
The diffusion model is trained for the task of deraining using the rainy image as a condition for generating rain-free results. Motivated by the power of diffusion models to learn the distribution of clean natural images, we train a diffusion model to learn the distribution of rain and, at the same time, make the model aware of the distribution of real-world rain-free images. For this, we first train a diffusion model for image generation using the ImageNet dataset, and we then train another diffusion model by ensuring that the weights of the model trained for deraining are aligned to the weights learned for real-world rain-free images.
Let $W_r$ denote the weights of the first model and $\theta$ denote the weights of the diffusion decoder estimated after each iteration through backpropagation. Then the weights of the second model are updated after each iteration of training according to, 
\begin{equation}
W' = \alpha.\theta + (1-\alpha).W_r,
\end{equation}
where $\alpha$ denotes the rate of Exponential Moving Average (EMA) for updating the decoder weights. The encoder weights are updated as such.

\begin{figure*}[ht!]
    \centering
    \resizebox{.9\linewidth}{!}{
    \setlength{\tabcolsep}{1pt}
    {\small
    \renewcommand{\arraystretch}{0.5} 
    \begin{tabular}{c c c c c c c c c}
    \captionsetup{type=figure, font=scriptsize}
    \raisebox{0.33in}{\rotatebox[origin=t]{90}{\scriptsize \emph{Input}}}&
    \includegraphics[width=0.12\linewidth]{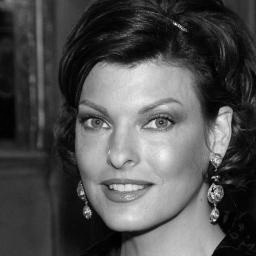}&
    \includegraphics[width=0.12\linewidth]{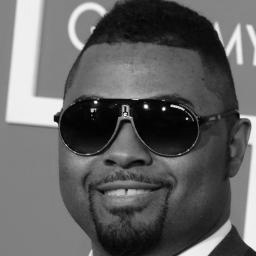}&
    \includegraphics[width=0.12\linewidth]{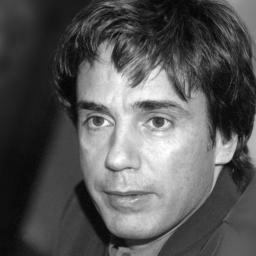}&
    \includegraphics[width=0.12\linewidth]{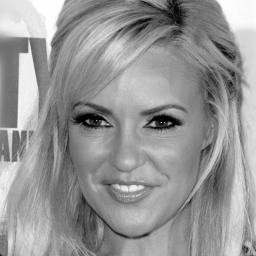}&
    \includegraphics[width=0.12\linewidth]{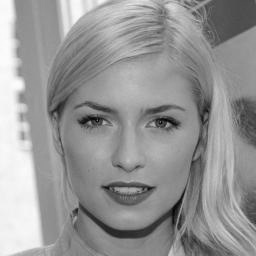}&
    \includegraphics[width=0.12\linewidth]{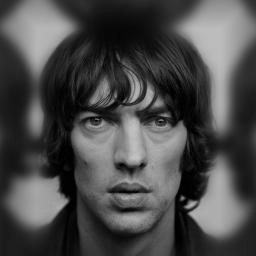}&
    \includegraphics[width=0.12\linewidth]{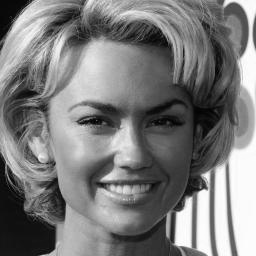}&
    \includegraphics[width=0.12\linewidth]{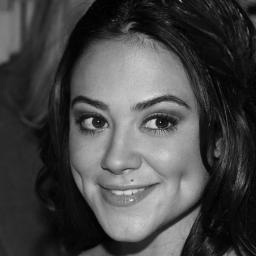}
    \tabularnewline
    \raisebox{0.33in}{\rotatebox[origin=t]{90}{\scriptsize \emph{ILVR Diffusion}}}&
    \includegraphics[width=0.12\linewidth]{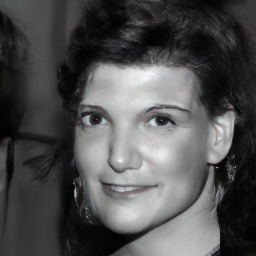}&
    \includegraphics[width=0.12\linewidth]{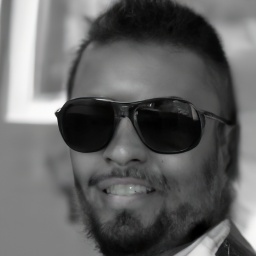}&
    \includegraphics[width=0.12\linewidth]{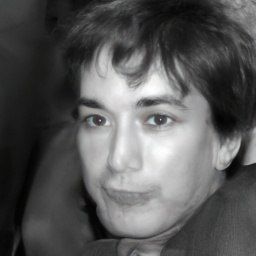}&
    \includegraphics[width=0.12\linewidth]{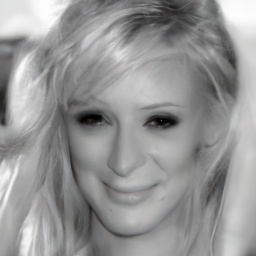}&
    \includegraphics[width=0.12\linewidth]{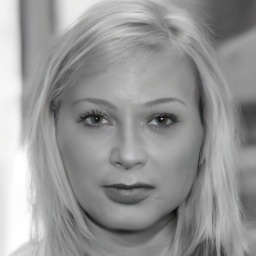}&
    \includegraphics[width=0.12\linewidth]{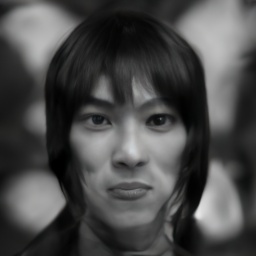}&
    \includegraphics[width=0.12\linewidth]{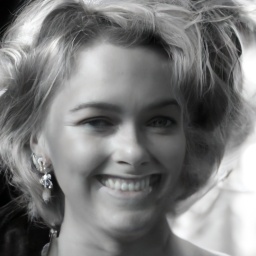}&
    \includegraphics[width=0.12\linewidth]{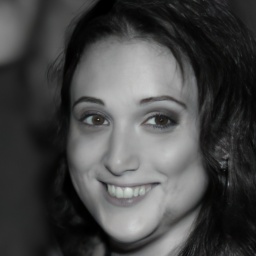}
    \tabularnewline
    \raisebox{0.33in}{\rotatebox[origin=t]{90}{\scriptsize \emph{SR3 Diffusion}}}&
    \includegraphics[width=0.12\linewidth]{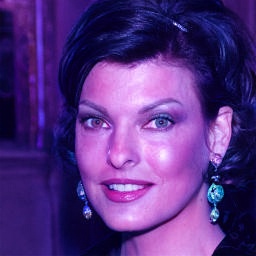}&
    \includegraphics[width=0.12\linewidth]{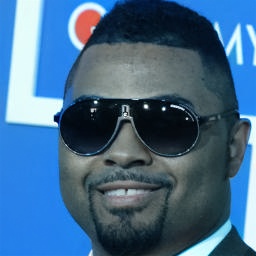}&
    \includegraphics[width=0.12\linewidth]{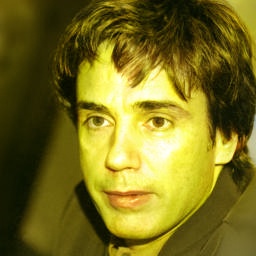}&
    \includegraphics[width=0.12\linewidth]{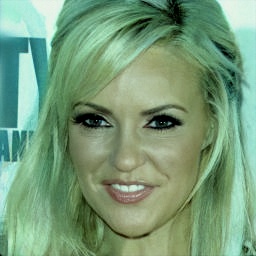}&
    \includegraphics[width=0.12\linewidth]{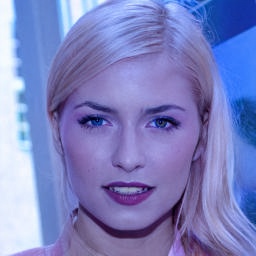}&
    \includegraphics[width=0.12\linewidth]{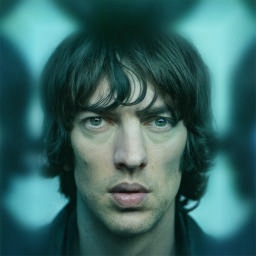}&
    \includegraphics[width=0.12\linewidth]{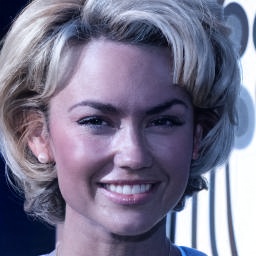}&
    \includegraphics[width=0.12\linewidth]{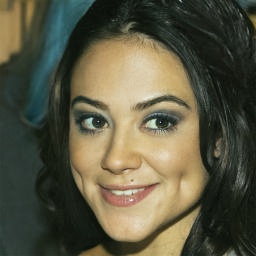}
    \tabularnewline
    \raisebox{0.33in}{\rotatebox[origin=t]{90}{\scriptsize \emph{Bi-noising (Ours)}}}&
    \includegraphics[width=0.12\linewidth]{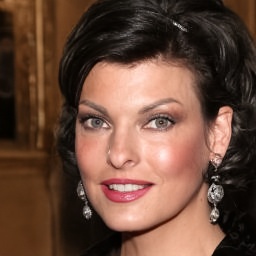}&
    \includegraphics[width=0.12\linewidth]{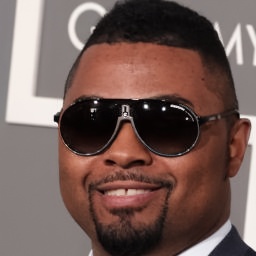}&
    \includegraphics[width=0.12\linewidth]{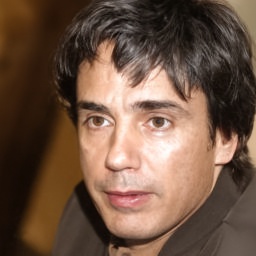}&
    \includegraphics[width=0.12\linewidth]{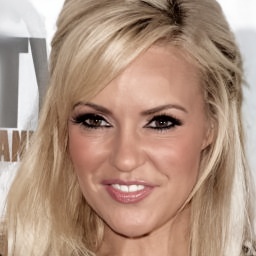}&
    \includegraphics[width=0.12\linewidth]{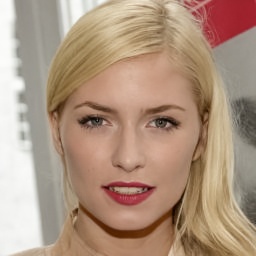}&
    \includegraphics[width=0.12\linewidth]{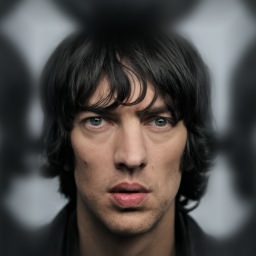}&
    \includegraphics[width=0.12\linewidth]{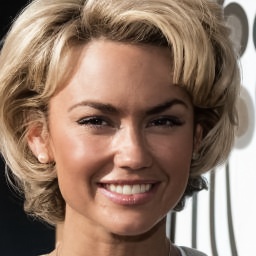}&
    \includegraphics[width=0.12\linewidth]{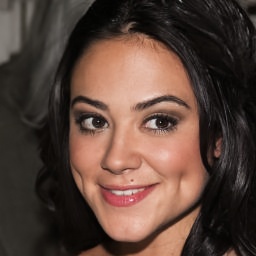}
    \tabularnewline
    \raisebox{0.33in}{\rotatebox[origin=t]{90}{\scriptsize \emph{Ground Truth}}}&
    \includegraphics[width=0.12\linewidth]{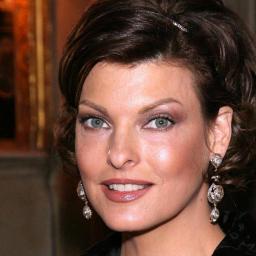}&
    \includegraphics[width=0.12\linewidth]{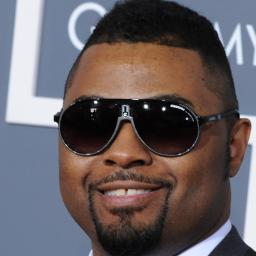}&
    \includegraphics[width=0.12\linewidth]{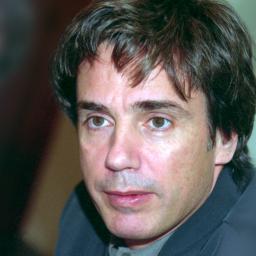}&
    \includegraphics[width=0.12\linewidth]{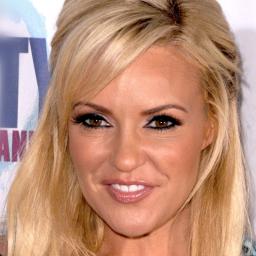}&
    \includegraphics[width=0.12\linewidth]{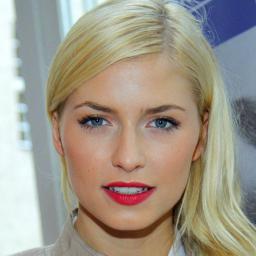}&
    \includegraphics[width=0.12\linewidth]{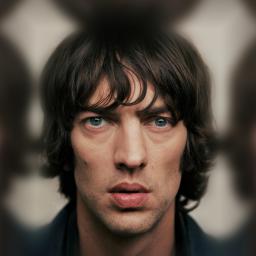}&
    \includegraphics[width=0.12\linewidth]{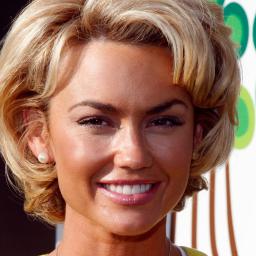}&
    \includegraphics[width=0.12\linewidth]{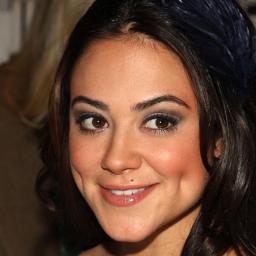}
    \end{tabular}
    }
    }
    \vspace{-.5\baselineskip}
    \caption{\textbf{Colorization visual result comparisons corresponding to the CelebAHQ dataset.}}
    \vspace{-.5\baselineskip}
\label{fig:colorization}
\end{figure*}

One observation from our experiments on image deraining while training by direct conditioning like in SR3 \cite{saharia2021image} was that the restored images suffered from artifacts and color channel shift which can be seen in Fig.~\ref{fig:rain800}. On further investigation, we found that this is due to incorrect conditioning of input during the training process. Specifically, for the task of image restoration with source-target pairs denoted as $(\bx_0,\by_0)$, existing methods optimize the weights of the network $\boldsymbol{\epsilon_\theta}(\cdot)$ modelling the reverse process of diffusion, by minimizing the $L_{simple}$ function defined in \cite{ho2020denoising} as
\begin{align}
    L_{\text {simple}}:=E_{t \sim[1, T], \epsilon \sim \mathcal{N}(0, \mathbf{I})}\left[\left\|\epsilon-\epsilon_{\theta}\left(\mathbf{x}_0, \mathbf{y}_t, t\right)\right\|^{2}\right]. 
\end{align}
% But training in this procedure, by conditioning on the same distorted image at each timestep, the weights for the initial layers of the denoising network are overfitted for the initial layers.This is because the input of the denoising network is the concatenated pair $(y_t,x_0)$. This causes a severe domain shift in the images for higher and lower values of $t$. To back up this claim, we provide illustrations in the supplementary paper. To correct this deviation, rather than conditioning on the input distorted image $x_0$ alone at each timestep $t$, we also condition on the estimated distorted image $x_t$ at time t according to, 
% \begin{align}
%   q(\bx_t | {\bx}_0) &:= \mathcal{N}(\bx_t | \sqrt{\bar \alpha_t} \tilde \bx_0, (1-\bar \alpha_t) \mathbf{I})\\
%   &= \sqrt{\bar{\alpha}_t} {\mathbf{x}}_0 + \epsilon \sqrt{1-\bar{\alpha}_{t}}, \epsilon \sim \mathcal{N}(0, \mathbf{I}).
% \end{align}
% We then modify the training objective as 
% \begin{align}
%     L_{\text {mod}}:=&\left\|\epsilon-\epsilon_{\theta}\left(\mathbf{x}_0, \mathbf{y}_t, t\right)\right\|^{2} \\
%     &+ D_\text{KL}(q(\mathbf{y}_{t-1} \vert \mathbf{y}_t, \mathbf{y}_0) \parallel p_\theta(\mathbf{y}_{t-1} \vert\mathbf{y}_t,\mathbf{x}_t)).
% \end{align}
% With this simple correction term, we modify the training at different timesteps to see the domain corrected $\bx_t$ along with $\bx_0$ hence leading to easier optimization of weights.

The training objective $L_{simple}$ holds the inherent assumption that during inference time $\by_t$, (i.e. the reconstructed image at time $t$ ) is close to the clean target. But for extreme cases where the intermediate diffusion outputs are not accurate during the initial steps of diffusion, the rain streaks continue to propagate through the diffusion process, as can be seen in Fig. \ref{fig:rain800}. This is because, inherently, the diffusion model works by predicting the noise present in $\by_t$ than the amount of degradation in it. To account for this, we add a correction prior $L_{corr}$ so that the network can give equally good output for high distortion levels. This term is defined by, 
 \begin{align}
    L_{\text {corr}}:=\alpha_t \left\|\epsilon_{\theta}\left(\mathbf{x}_t, \mathbf{x}_0, t\right)-\epsilon_{\theta}\left(\mathbf{y}_t, \mathbf{x}_0, t\right)\right\|^{2}.
\end{align}
The final objective for training the network is,
 \begin{align}
    L_{\text {final}}=L_{\text {simple}}+\lambda_{corr}L_{\text {corr}}.
\end{align}
The value of $\lambda_{corr}$ is empirically set equal to $0.001$ for all experiments.

\section{Experiments}
To demonstrate the restoration capacity of our method, we evaluate our method with several experimental settings following the most representative diffusion models, \ie, ILVR~\cite{choi2021ilvr} and SR3~\cite{saharia2021image} based on the Guided-diffusion architecture~\cite{nichol2021glide}.
Following the common practice that pixel-wise metrics, \ie, PSNR and SSIM cannot comprehensively denote the visual quality of restored results, we utilize FID
% \footnote{\url{https://github.com/mseitzer/pytorch-fid}}
and LPIPS as the additional metrics for evaluation. The tasks in which we evaluate our method on are
\begin{itemize}[noitemsep]
  \item Conditional image restoration which is trained on the FFHQ~\cite{russakovsky2015imagenet} dataset (70000 images) and evaluated on the CelebA-HQ~\cite{liu2015deep,karras2017progressive} dataset (first 3000 images) with a resolution of $256\times 256$ pixels.
  
  \item Conditional image restoration which is $4\times$ face super-resolution trained on the FFHQ~\cite{russakovsky2015imagenet} dataset and evaluated on the CelebA-HQ~\cite{liu2015deep,karras2017progressive} dataset (first 3000 images) with a resolution of $256\times 256$ pixels.
  
  % \item Natural image restoration which is $4\times$ super-resolution with noise $\sigma=0.05$ on the ImageNet 1K~\cite{russakovsky2015imagenet} dataset, and it is evaluated on the corresponding dev-split (the first image of each classes). The ground truth label is applied as the class condition for sampling.
  
  \item Image turbulence removal follows the turbulence simulation settings~\cite{nair2022ddpm} on the FFHQ dataset and conducts evaluation on the real long-range imaging images~\cite{miller2019data}.

 \item Image deraining which is conducted on the Rain800~\cite{zhang2019image} dataset and Jorder 200L~\cite{yang2017deep} dataset with their respective train sets. The diffusion models conduct in a resolution of $256\times 256$ pixels.

\end{itemize}

Note that for the first three tasks, the diffusion models are trained on the FFHQ dataset for face generation.
For the last task, the diffusion models are trained on the ImageNet dataset for natural image generation. The unconditional model utilized has never seen the validation dataset during its training process for all of these cases.

\subsection{Colorization}

Colorization aims at reconstructing grayscale images with colors that are fitted to natural statistics and image semantics.
The grayscale image is obtained by averaging the values at red, green, and blue channels of the corresponding colour image.
We empirically observed that conditional denoising diffusion models fail at colorization. Even though they can preserve the fine-grained details, unnatural colors always exist in their reconstructed results.
In contrast, the method that adopts our proposed bi-noising diffusion is capable of correcting the reconstruction with more semantics and accurate color descriptions.
The quantitative performance comparison is shown in Tab.~\ref{tab:colorizarion_performance}, where our method achieves 7.906dB higher PSNR than the one without pretraining.
The visual results in Fig.~\ref{fig:colorization} further clarify the improvements that come from more globally consistent colors and tones of our results, even though the pretraining had never seen the ground truth before.
In contrast, a similar method, i.e., ILVR cannot deal with the colorization task even though it also utilizes a pretrained unconditional model, which demonstrates the superiority of our proposed DDRP in such tasks. Therefore, we argue that utilizing the priors plays a crucial role in ensuring the color naturalism.

\begin{table}[ht!]
  \centering
\resizebox{1.0\linewidth}{!}{
  \begin{tabular}{lcccc}
    \toprule
    ~Method &~PSNR $\uparrow$~&~SSIM $\uparrow$~&~FID $\downarrow$~&~LPIPS $\downarrow$~\\
    \midrule
    % ~Stochastic Generation & - & - & 22.40 & - \\
    % \\
    ILVR Diffusion~\cite{choi2021ilvr} & 18.3936 & 0.5674 & 86.2642 & 0.5008 \\
    SR3 Diffusion~\cite{saharia2021image} & \emph{\underline{19.1647}} & \emph{\underline{0.8680}} & \emph{\underline{13.8126}} & \emph{\underline{0.2959}} \\
    Bi-Noising (Ours) & \textbf{27.0707} & \textbf{0.9531} & \textbf{12.6796} & \textbf{0.1417} \\
    \bottomrule
  \end{tabular}}
  \vspace{-.1in}
  \caption{\textbf{Colorization results corresponding to the CelebAHQ dataset.} The best and second-best performace is indicated with \textbf{blod} and \emph{\underline{italic}} respectively. We use $\uparrow$ and $\downarrow$ to suggest high/lower score should be achieved by better methods.}
  \label{tab:colorizarion_performance}
  \vspace{-.1in}
\end{table}

\subsection{Face Super-resolution}
Face super-resolution is the other representative task in image restoration, and it is widely evaluated in the other denoising diffusion-based restoration works.
We follow the experimental settings of SR3 and ILVR, i.e., restore $256\times 256$ face images from $64\times 64$ face images downsampled by Bicubic interpolation.
The implementation details of PULSE~\cite{menon2020pulse}, ILVR~\cite{choi2021ilvr}, SR3~\cite{saharia2021image} are presented in the supplementary file.
From Fig.~\ref{fig:facesr}, one can notice that our method achieves the best visual quality compared with the other methods.
Compared with the state-of-the-art face super-resolution method based on GAN priors, our method better preserves the identity of the restored face images.
As can be seen from Tab.~\ref{tab:sr-performance}, our  method significantly outperforms the other methods in terms of the distortion measures, i.e., PSNR and SSIM with 4.8316 dB and 0.04 better than the second one.
Though our results in the FID metric are not better than ILVR, FID doesn't denote the reconstruction accuracy that is curcial for super-resolution.
Therefore, the above results demonstrate our performance superiority.

\begin{figure}[t]
    \centering
    \resizebox{\linewidth}{!}{
    \setlength{\tabcolsep}{1pt}
    {\small
    \renewcommand{\arraystretch}{0.5} 
    \begin{tabular}{c c c c c}
    \captionsetup{type=figure, font=scriptsize}
    \raisebox{0.3in}{\rotatebox[origin=t]{90}{\scriptsize \emph{Input}}}&
    \includegraphics[width=0.22\linewidth]{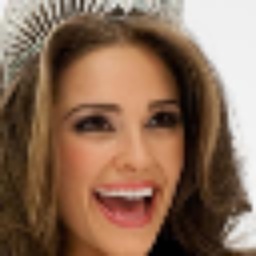} &
    \includegraphics[width=0.22\linewidth]{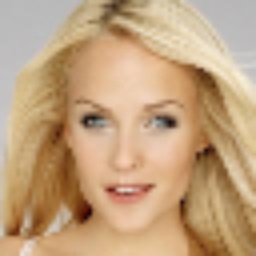} &
    \includegraphics[width=0.22\linewidth]{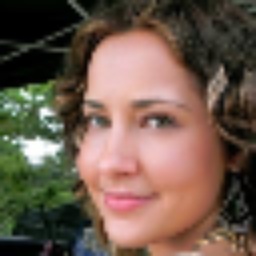} &
    \includegraphics[width=0.22\linewidth]{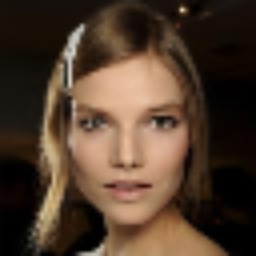}
    \tabularnewline
    \raisebox{0.3in}{\rotatebox[origin=t]{90}{\scriptsize \emph{PULSE}}}&
    \includegraphics[width=0.22\linewidth]{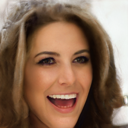} &
    \includegraphics[width=0.22\linewidth]{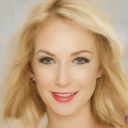} &
    \includegraphics[width=0.22\linewidth]{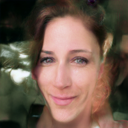} &
    \includegraphics[width=0.22\linewidth]{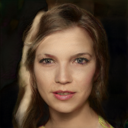}
    \tabularnewline
    \raisebox{0.3in}{\rotatebox[origin=t]{90}{\scriptsize \emph{ILVR Diffusion}}}&
    \includegraphics[width=0.22\linewidth]{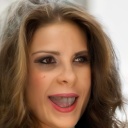} &
    \includegraphics[width=0.22\linewidth]{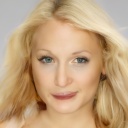} &
    \includegraphics[width=0.22\linewidth]{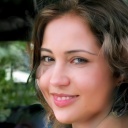} &
    \includegraphics[width=0.22\linewidth]{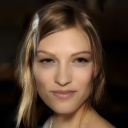}
    \tabularnewline
    \raisebox{0.3in}{\rotatebox[origin=t]{90}{\scriptsize \emph{SR3 Diffusion}}}&
    \includegraphics[width=0.22\linewidth]{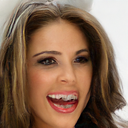} &
    \includegraphics[width=0.22\linewidth]{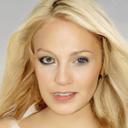} &
    \includegraphics[width=0.22\linewidth]{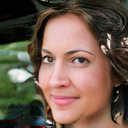} &
    \includegraphics[width=0.22\linewidth]{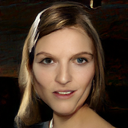}
    \tabularnewline
    \raisebox{0.3in}{\rotatebox[origin=t]{90}{\scriptsize \emph{Bi-noising (Ours)}}}&
    \includegraphics[width=0.22\linewidth]{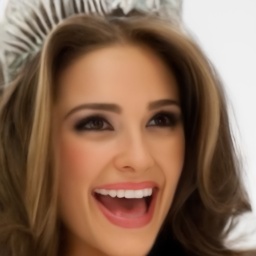} &
    \includegraphics[width=0.22\linewidth]{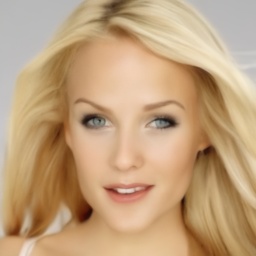} &
    \includegraphics[width=0.22\linewidth]{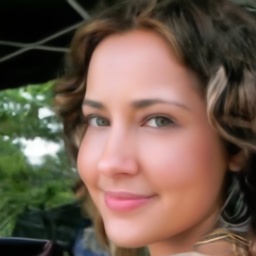} &
    \includegraphics[width=0.22\linewidth]{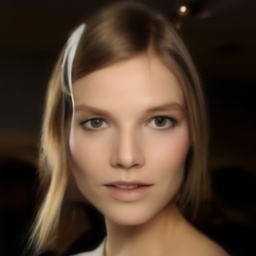}
    \tabularnewline
    \raisebox{0.3in}{\rotatebox[origin=t]{90}{\scriptsize \emph{Ground Truth}}}&
    \includegraphics[width=0.22\linewidth]{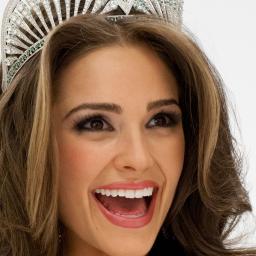} &
    \includegraphics[width=0.22\linewidth]{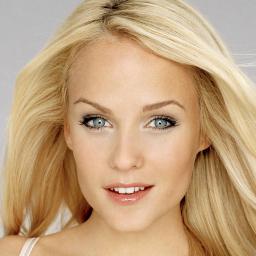} &
    \includegraphics[width=0.22\linewidth]{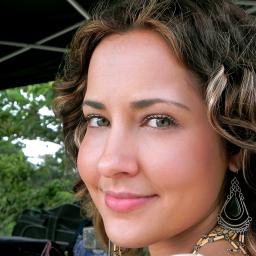} &
    \includegraphics[width=0.22\linewidth]{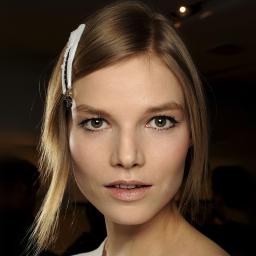}    
    \end{tabular}
    }
    }
    \vspace{-1\baselineskip}
    \caption{\textbf{$4\times$ super-resolution visual result comparisons corresponding to the CelebAHQ dataset.}}
    \label{fig:facesr}
\end{figure}

\begin{table}[htbp]
  \centering
    \resizebox{1.0\linewidth}{!}{
  \begin{tabular}{lcccc}
    \toprule
    Method &~PSNR $\uparrow$~&~SSIM $\uparrow$~&~FID $\downarrow$~&~LPIPS $\downarrow$~\\
    \midrule
    % Stochastic Generation & - & - & 22.40 & - \\
    PULSE~\cite{menon2020pulse} & \emph{\underline{23.5769}} & \emph{\underline{0.6794}} & 31.2309 & 0.3832 \\
    ILVR Diffusion ~\cite{choi2021ilvr} & 22.5374 & 0.6150 & \textbf{20.4621} & 0.3393 \\
    SR3 Diffusion~\cite{saharia2021image} & 22.8290 & 0.6442 & 29.8932 & \emph{\underline{0.3350}} \\
    Bi-Noising (Ours) & \textbf{29.3996} & \textbf{0.8414} & \emph{\underline{24.5632}} & \textbf{0.1809} \\
    \bottomrule
  \end{tabular}}
    \vspace{-.5\baselineskip}
  \caption{\textbf{$4\times$ super-resolution results corresponding to the CelebAHQ dataset.}}
  \label{tab:sr-performance}
\end{table}

\begin{figure*}[t]
    \centering
    \resizebox{.92\linewidth}{!}{
    \setlength{\tabcolsep}{1pt}
    {\small
    \renewcommand{\arraystretch}{0.5} 
    \begin{tabular}{c c c c c c c c c}
    \captionsetup{type=figure, font=small}
    \raisebox{0.28in}{\rotatebox[origin=t]{90}{\scriptsize \emph{Input}}}&
    \includegraphics[width=0.16\linewidth, height = 0.10\linewidth]{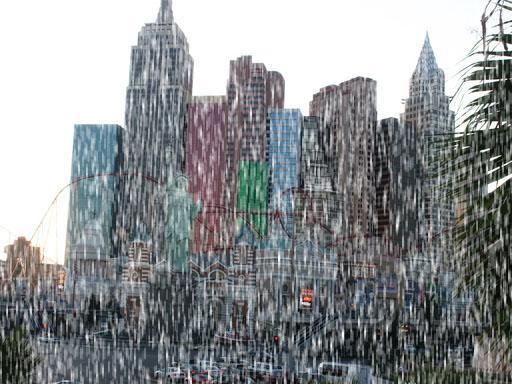}&
    \includegraphics[width=0.16\linewidth]{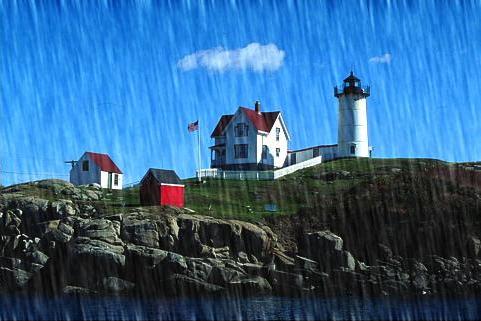}&
    \includegraphics[width=0.16\linewidth]{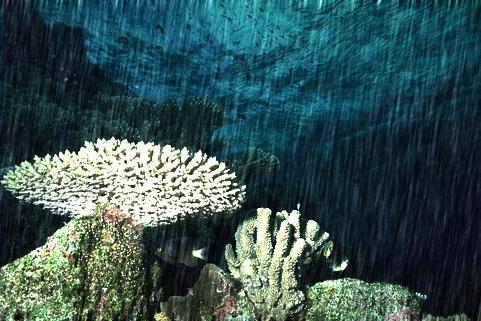}&
    \includegraphics[width=0.16\linewidth]{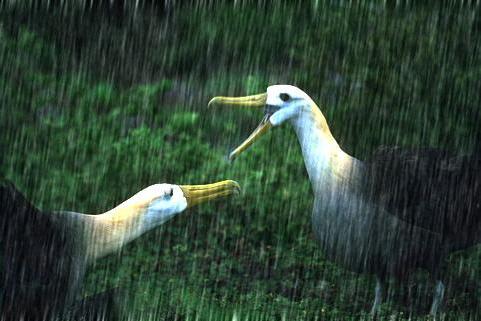}&
    \includegraphics[width=0.16\linewidth]{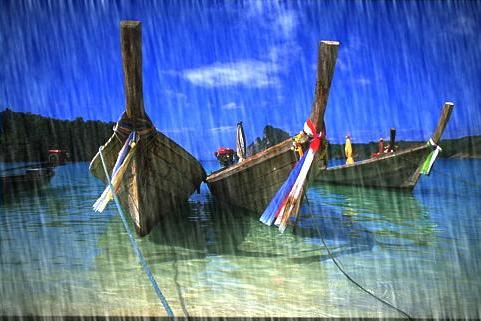}&
    \includegraphics[width=0.16\linewidth]{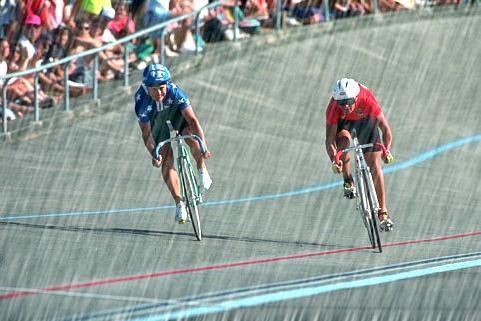}
    \tabularnewline
    \raisebox{0.32in}{\rotatebox[origin=t]{90}{\scriptsize \emph{ILVR Difusion}}}&
    \includegraphics[width=0.16\linewidth, height = 0.105\linewidth]{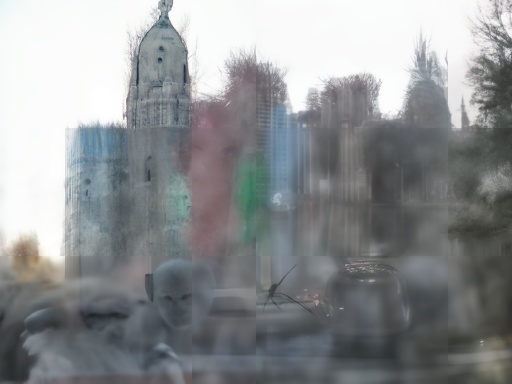}&
    \includegraphics[width=0.16\linewidth]{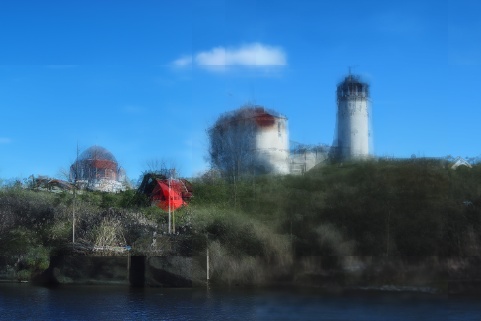}&
    \includegraphics[width=0.16\linewidth]{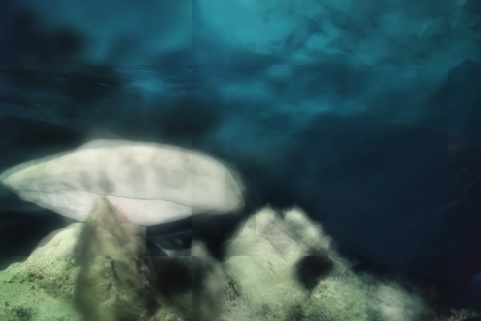}&
    \includegraphics[width=0.16\linewidth]{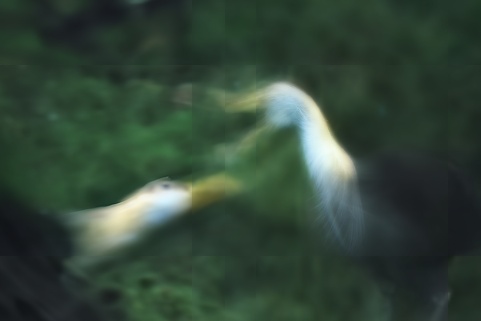}&
    \includegraphics[width=0.16\linewidth]{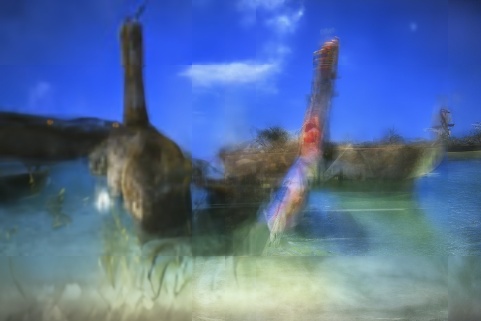}&
    \includegraphics[width=0.16\linewidth]{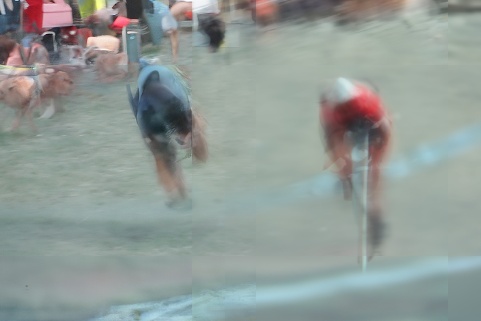}
    \tabularnewline
    \raisebox{0.28in}{\rotatebox[origin=t]{90}{{\scriptsize \emph{SR3 Diffusion}}}}&
     \includegraphics[width=0.16\linewidth, height = 0.105\linewidth]{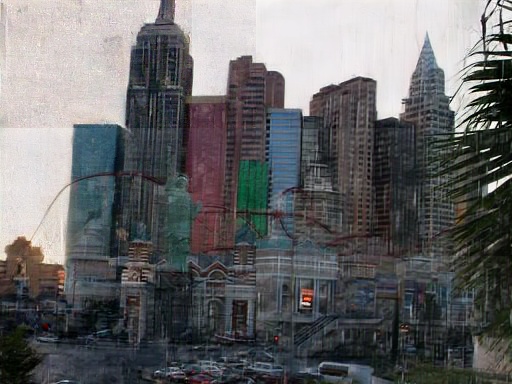}&
    \includegraphics[width=0.16\linewidth]{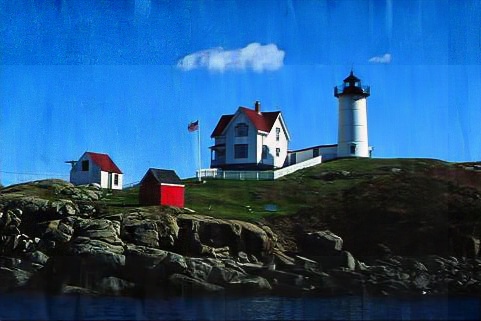}&
    \includegraphics[width=0.16\linewidth]{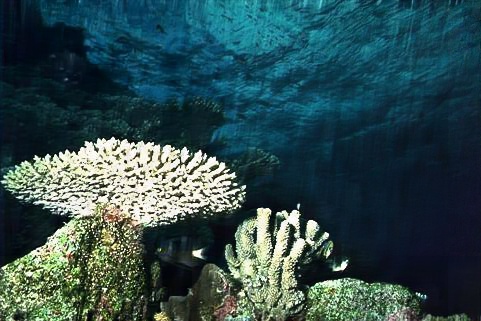}&
    \includegraphics[width=0.16\linewidth]{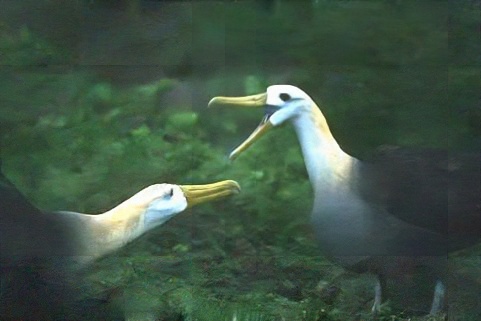}&
    \includegraphics[width=0.16\linewidth]{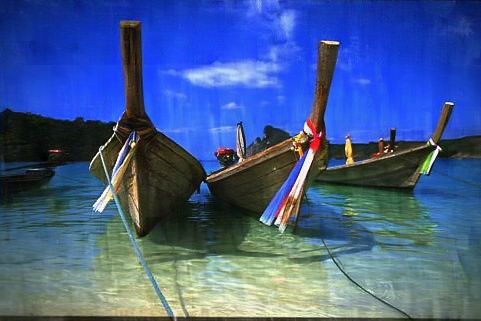}&
    \includegraphics[width=0.16\linewidth]{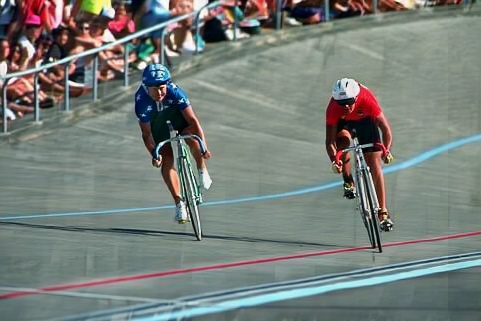}
    \tabularnewline
    \raisebox{0.3in}{\rotatebox[origin=t]{90}{\scriptsize \emph{Bi-Noising (Ours)}}}&
    \includegraphics[width=0.16\linewidth, height = 0.105\linewidth]{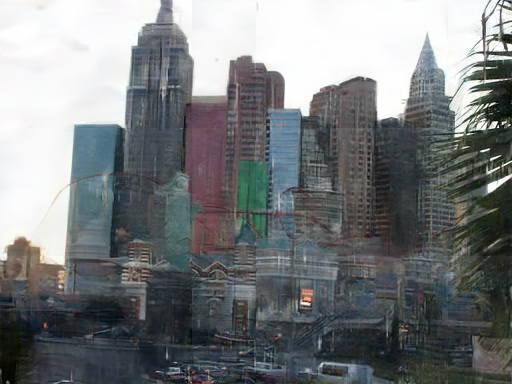}&
    \includegraphics[width=0.16\linewidth]{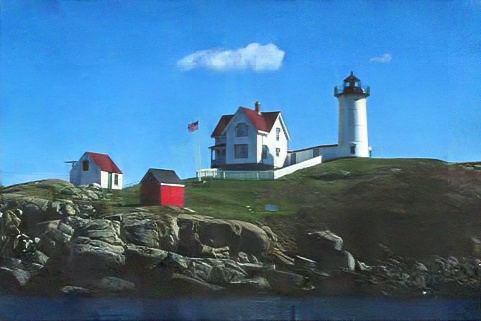}&
    \includegraphics[width=0.16\linewidth]{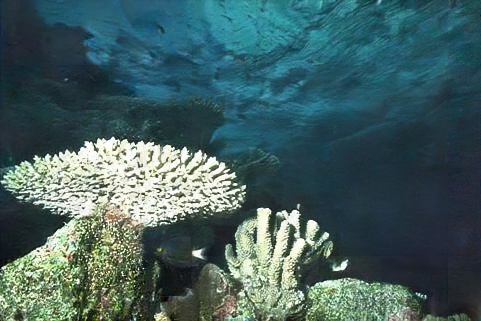}&
    \includegraphics[width=0.16\linewidth]{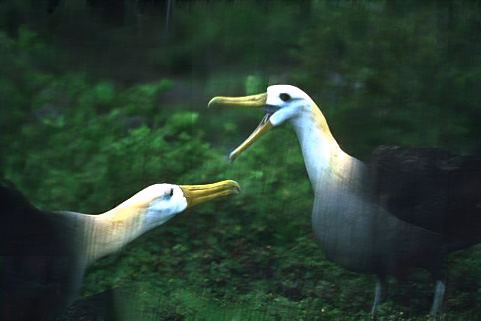}&
    \includegraphics[width=0.16\linewidth]{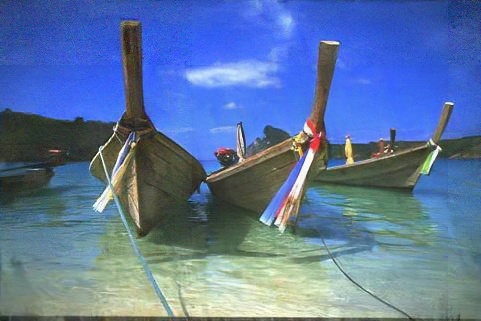}&
    \includegraphics[width=0.16\linewidth]{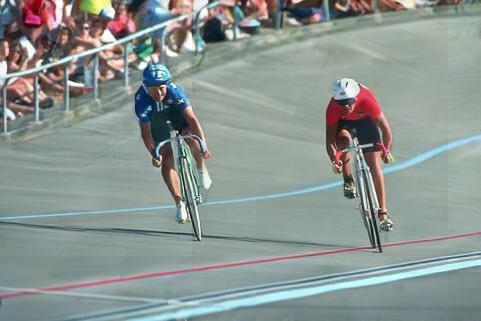}
    \tabularnewline
    \raisebox{0.28in}{\rotatebox[origin=t]{90}{\scriptsize \emph{Ground Truth}}}&
     \includegraphics[width=0.16\linewidth, height = 0.10\linewidth]{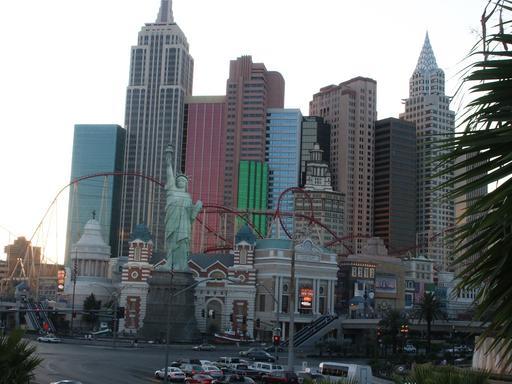}&
    \includegraphics[width=0.16\linewidth]{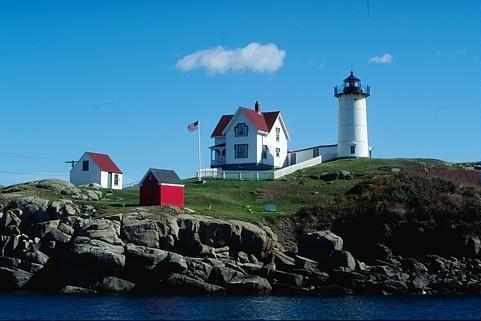}&
    \includegraphics[width=0.16\linewidth]{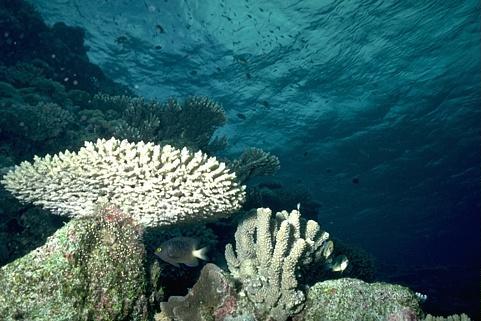}&
    \includegraphics[width=0.16\linewidth]{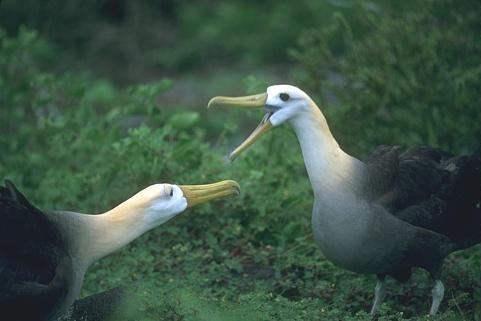}&
    \includegraphics[width=0.16\linewidth]{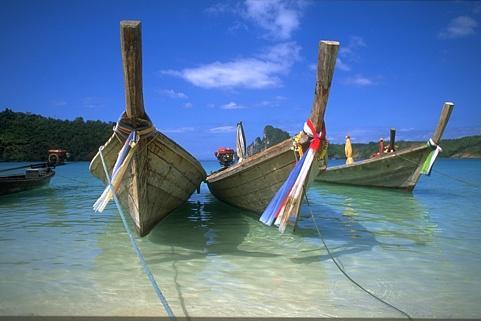}&
    \includegraphics[width=0.16\linewidth]{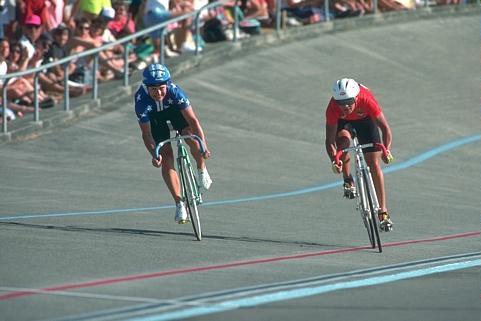}
    \end{tabular}
    }
    }
    \vspace{-.5\baselineskip}
    \caption{\textbf{Deraining visual result comparisons corresponding to the Rain 800 dataset.}}
    \label{fig:rain800}
    \vspace{-.5\baselineskip}
\end{figure*}

\begin{figure}[t]
    \centering
    \resizebox{\linewidth}{!}{
    \setlength{\tabcolsep}{1pt}
    {\small
    \renewcommand{\arraystretch}{0.5} 
    \begin{tabular}{c c c c c c c c c}
    \captionsetup{type=figure, font=small}
    \raisebox{0.28in}{\rotatebox[origin=t]{90}{\scriptsize \emph{Input}}}&
    \includegraphics[width=.24\linewidth, height=.2\linewidth]{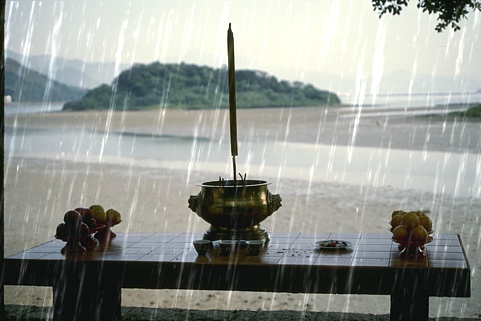}&
    \includegraphics[width=.24\linewidth, height=.2\linewidth]{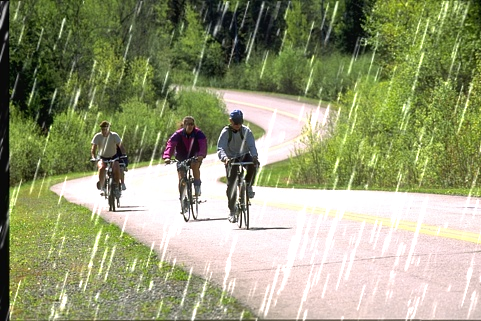}&
    \includegraphics[width=.24\linewidth, height=.2\linewidth]{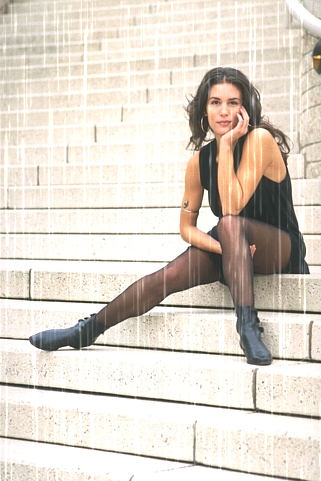}&
    \includegraphics[width=.24\linewidth, height=.2\linewidth]{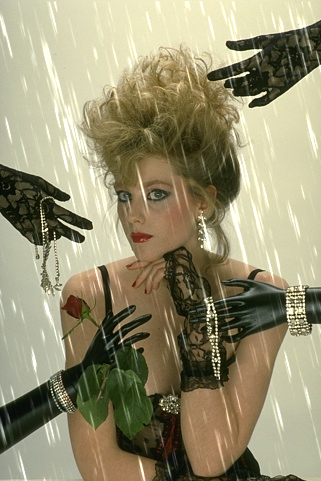}&
    \tabularnewline
    \raisebox{0.32in}{\rotatebox[origin=t]{90}{\scriptsize \emph{ILVR Difusion}}}&
    \includegraphics[width=.24\linewidth, height=.2\linewidth]{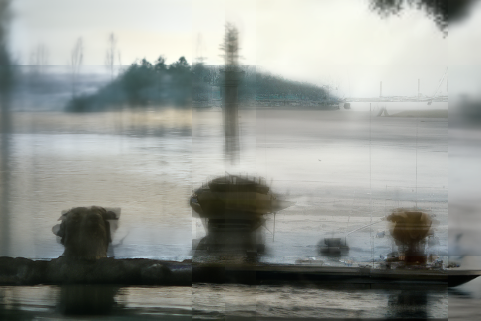}&
    \includegraphics[width=.24\linewidth, height=.2\linewidth]{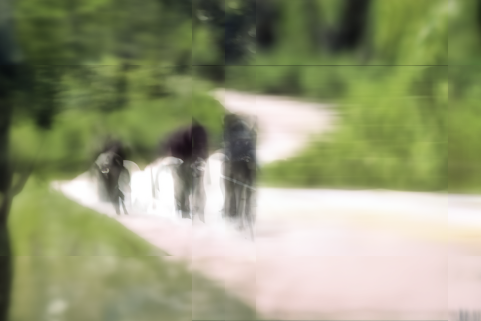}&
    \includegraphics[width=.24\linewidth, height=.2\linewidth]{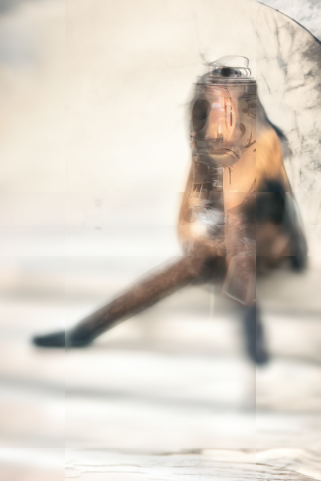}&
    \includegraphics[width=.24\linewidth, height=.2\linewidth]{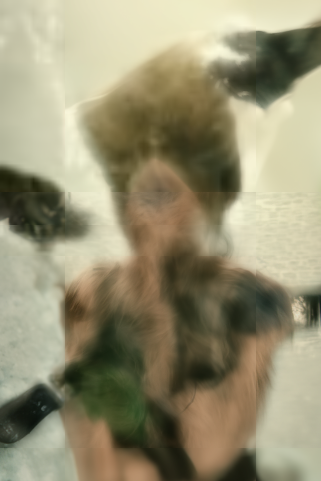}&
    \tabularnewline
    \raisebox{0.32in}{\rotatebox[origin=t]{90}{\scriptsize \emph{SR3 Difusion}}}&
    \includegraphics[width=.24\linewidth, height=.2\linewidth]{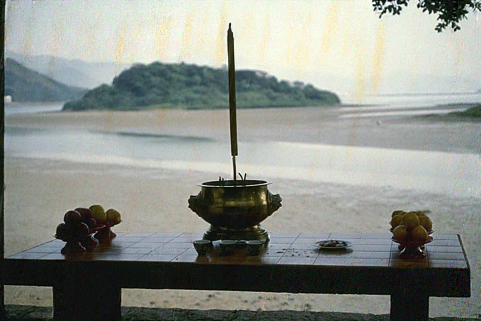}&
    \includegraphics[width=.24\linewidth, height=.2\linewidth]{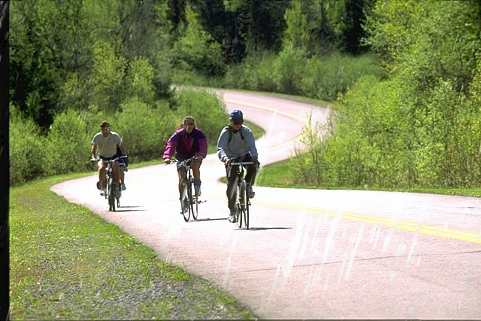}&
    \includegraphics[width=.24\linewidth, height=.2\linewidth]{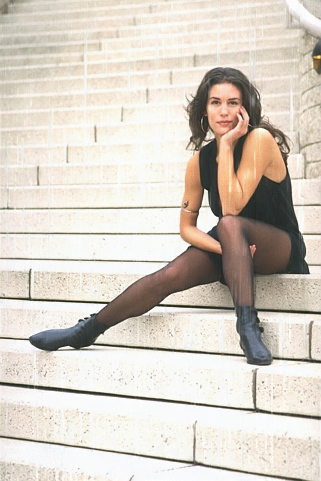}&
    \includegraphics[width=.24\linewidth, height=.2\linewidth]{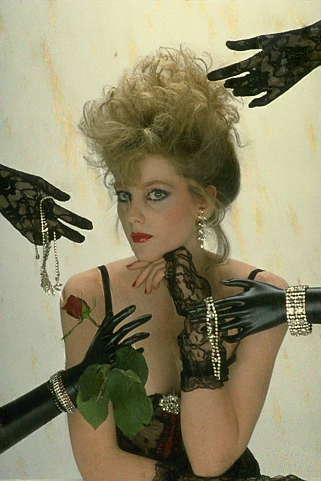}&
    \tabularnewline
    \raisebox{0.3in}{\rotatebox[origin=t]{90}{\scriptsize \emph{Bi-Noising (Ours)}}}&
    \includegraphics[width=.24\linewidth, height=.2\linewidth]{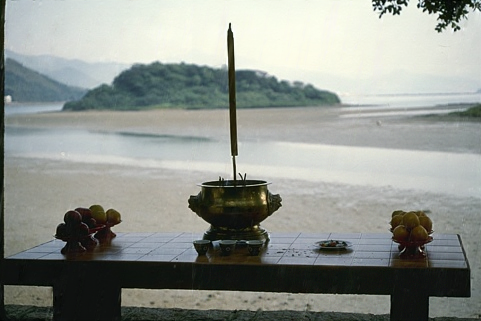}&
    \includegraphics[width=.24\linewidth, height=.2\linewidth]{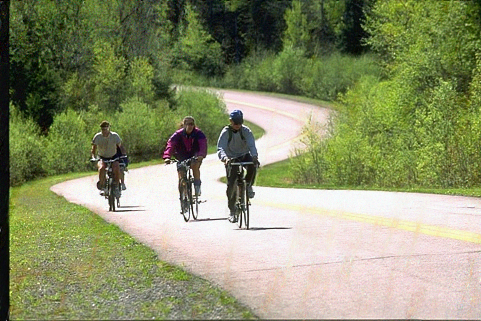}&
    \includegraphics[width=.24\linewidth, height=.2\linewidth]{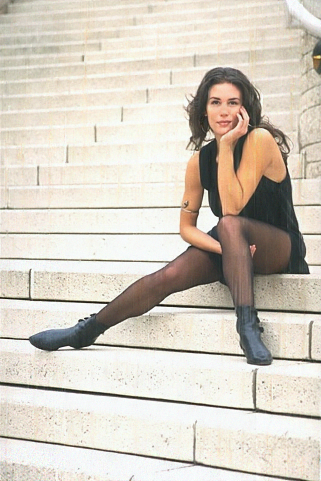}&
    \includegraphics[width=.24\linewidth, height=.2\linewidth]{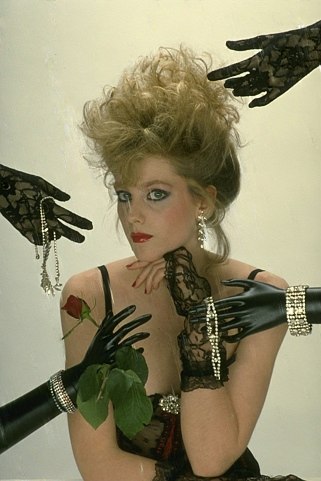}&
    \tabularnewline
    \raisebox{0.28in}{\rotatebox[origin=t]{90}{\scriptsize \emph{Ground Truth}}}&
    \includegraphics[width=.24\linewidth, height=.2\linewidth]{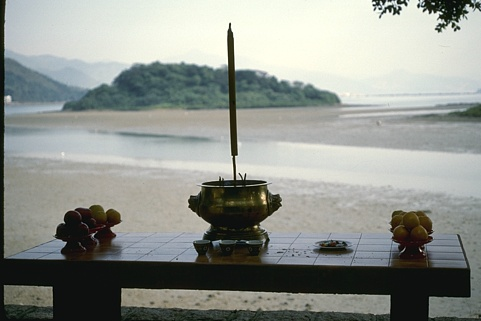}&
    \includegraphics[width=.24\linewidth, height=.2\linewidth]{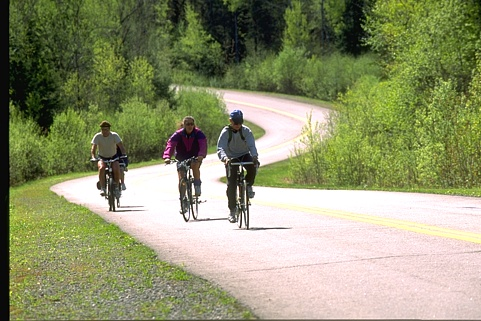}&
    \includegraphics[width=.24\linewidth, height=.2\linewidth]{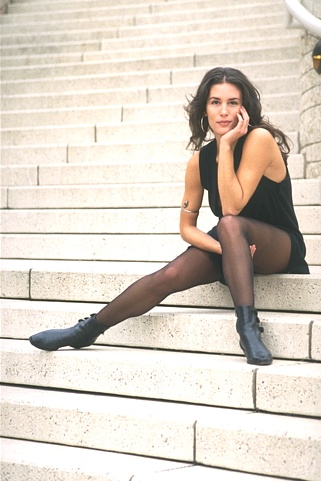}&
    \includegraphics[width=.24\linewidth, height=.2\linewidth]{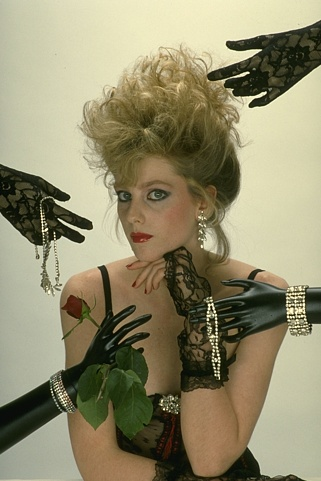}&
    \end{tabular}
    }
    }
    \vspace{-.5\baselineskip}
    \caption{\textbf{Deraining visual result comparisons corresponding to the JORDER-200 dataset.}}
    \label{fig:jorder200}
    \vspace{-1.5\baselineskip}
\end{figure}

\subsection{Image Deraining}
We perform single image deraining on two popular draining datasets. Namely, the Jorder 200L dataset which contains large rain streaks, and the Rain800 dataset which contains realistic rain. Since no diffusion-based deraining method has been proposed in literature before, we perform comparisons after retraining the models proposed for super-resolution in the  literature. Specifically, we perform comparisons with ILVR diffusion~\cite{choi2021ilvr} and conditional diffusion models, and we include the improvements brought about by our modules. To evaluate the reconstruction quality, we use the PSNR and SSIM metrics. To assess the quality of images produced by various  methods, we use LPIPS and NIQE as metrics. 
As we can see from Tab.~\ref{tab:j200}, the proposed conditioning loss functions bring significant improvement for all metrics in the JORDER 200L dataset \cite{yang2017deep}, obtaining about $2.45$ dB PSNR over the exiting method as well as giving realistic natural images.
The visual comparisons in Fig.~\ref{fig:rain800} and Fig.~\ref{fig:jorder200} further demonstrate our method on the visual quality compared with the other methods.

\begin{table}[htbp]
  \centering
  \resizebox{1.0\linewidth}{!}{
  \begin{tabular}{lcccc}
    \toprule
    & \multicolumn{4}{c}{Jorder 200L dataset} \\
    \cmidrule(r){2-5}
    Method & PSNR $\uparrow$  &SSIM $\uparrow$ & LPIPS $\downarrow$ & NIQE $\downarrow$ \\
    \midrule
    Rain Images (Input) & $26.70$ &$0.8439$&$0.2411$ &$4.131$ \\
    ILVR Diffusion~\cite{choi2021ilvr} & $21.22$ & $0.4942$ & \textbf{0.0972} & $6.467$ \\
    SR3 Diffusion~\cite{saharia2021image} & \emph{\underline{31.45}} & \emph{\underline{0.9091}} & \emph{\underline{0.1779}} & \emph{\underline{3.588}}  \\
    Bi-Noising (Ours) & \textbf{33.90} & \textbf{0.9555} & \textbf{0.0972} & \textbf{3.232}  \\
    \bottomrule
    \end{tabular}}
    \vspace{-.5\baselineskip}
    \caption{\textbf{Restoration results comparison on the Jorder 200L dataset with the other re-trained diffusion models.}}
    \label{tab:j200}
    \vspace{-.5\baselineskip}
\end{table}

\begin{figure*}[ht!]
 \centering
 \begin{minipage}[c]{.73\linewidth}
 \resizebox{\linewidth}{!}{
  \begin{tabular}{rccccccc}
    \toprule
    & \multicolumn{4}{c}{Methods} & \multicolumn{3}{c}{Ours}  \\
    \cmidrule(r){2-5} \cmidrule(r){6-8} ~Settings & Ho et al.~\cite{ho2020denoising} & Dhariwal et al.~\cite{dhariwal2021diffusion} & Nichol et al.~\cite{nichol2021glide} & Ho et al.~\cite{ho2022classifier}  & w/o parametric & w/o full guidance & Bi-Noising \\
    \midrule
    classifier guidance~\cite{dhariwal2021diffusion} & & \checkmark \\
    CLIP guidance~\cite{nichol2021glide} & & & \checkmark \\
    classifier-free guidance~\cite{ho2022classifier} & & & & \checkmark & \checkmark \\
    alternative guidance & & & & & & \checkmark  \\
    Bi-Noising & & & & & \checkmark & \checkmark & \checkmark \\
    \midrule
    PSNR $\uparrow$ & 19.16 & 20.10 & 23.14 & 25.91 & 26.46 & \emph{\underline{26.81}} & \textbf{27.07} \\
    Parameters (M) $\downarrow$ & \textbf{93.6} & 147.7 & 243.2 & \textbf{93.6} & \textbf{93.6} & 187.2 & 187.2     \\
    Running Time (s) $\downarrow$ & \textbf{1.6} & 4.9 & 3.4 & 3.1 & 3.1 & \emph{\underline{2.3}} & 3.1 \\
    \bottomrule
  \end{tabular}
  }
\vspace{-.5\baselineskip}
\captionof{table}{\textbf{Result comparisons between different prior parameterizations.}}
  \label{tab:param}
  \end{minipage}
  \begin{minipage}[c]{.26\linewidth}
  \includegraphics[width=\linewidth]{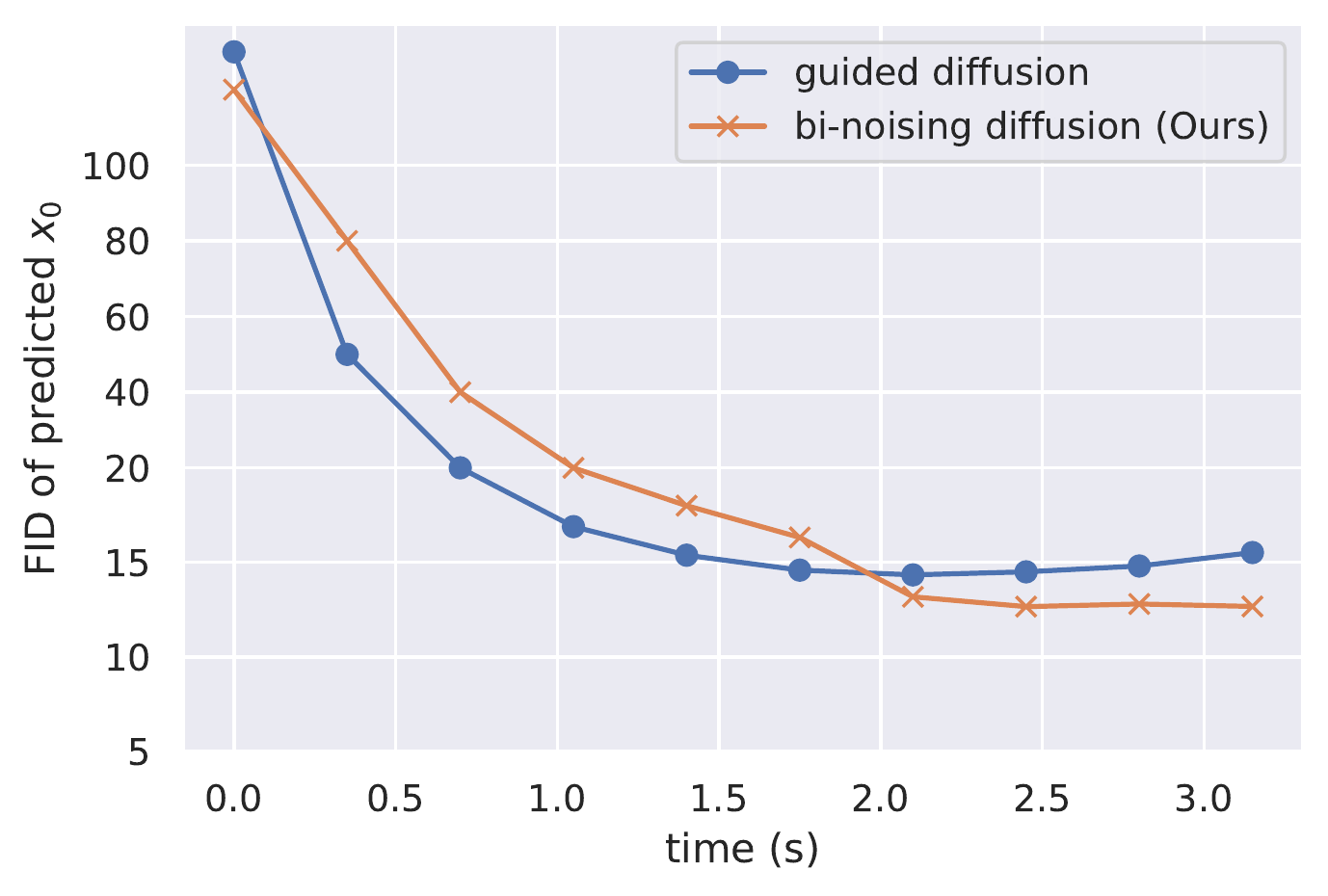}
\vspace{-1.8\baselineskip}
  \captionof{figure}{\textbf{FID v.s. Time.}}
  \label{fig:fid}
  \end{minipage}
\vspace{-1\baselineskip}
\end{figure*}

% \begin{table}[htbp]
%   \centering
%   \resizebox{1.0\linewidth}{!}{
%   \begin{tabular}{lcccc}
%     \toprule
%     & \multicolumn{4}{c}{Rain800 dataset} \\
%     \cmidrule(r){2-5}
%     Method & PSNR $\uparrow$  &SSIM $\uparrow$ & LPIPS $\downarrow$ & NIQE $\downarrow$ \\
%     \midrule
%     Rain Images (Input) &\emph{\underline{22.45}} &$0.6796$& $0.3815$ & $5.283$ \\
%     ILVR Diffusion~\cite{choi2021ilvr} & $19.65$ & $0.5238$ & $0.5478$& $5.564$  \\
%     SR3 Diffusion~\cite{saharia2021image} & \emph{\underline{22.45}} & \textbf{0.8185} & \textbf{0.2869} & \textbf{3.307}   \\
%     Bi-Noising (Ours) & \textbf{24.28} & \textbf{0.8068} & \emph{\underline{0.3089}} & \emph{\underline{3.339}}  \\
%     \bottomrule
%     \end{tabular}}
%     \vspace{-.1in}
%     \caption{\textbf{Restoration results comparison on the Rain800 dataset with the other re-trained diffusion models.}}
%     \label{tab:j2001}
% \end{table}

\subsection{Turbulence Removal}
\begin{figure}[htbp]
    \centering
    \resizebox{\linewidth}{!}{
    \setlength{\tabcolsep}{1pt}
    {\small
    \renewcommand{\arraystretch}{0.5} 
    \begin{tabular}{c c c c c}
    \captionsetup{type=figure, font=scriptsize}
    \raisebox{0.3in}{\rotatebox[origin=t]{90}{\scriptsize \emph{Input}}}&
    \includegraphics[width=0.22\linewidth]{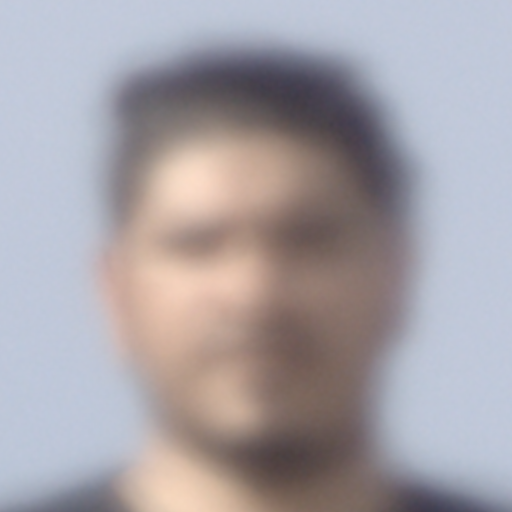} &
    \includegraphics[width=0.22\linewidth]{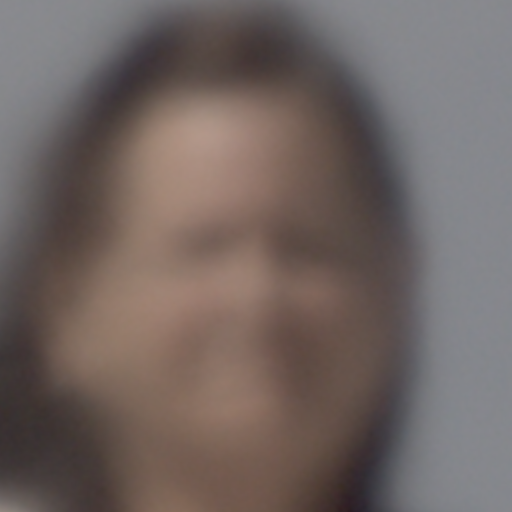} &
    \includegraphics[width=0.22\linewidth]{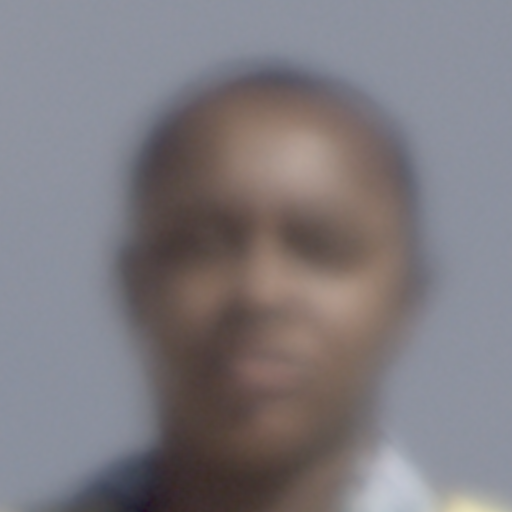} &
    \includegraphics[width=0.22\linewidth]{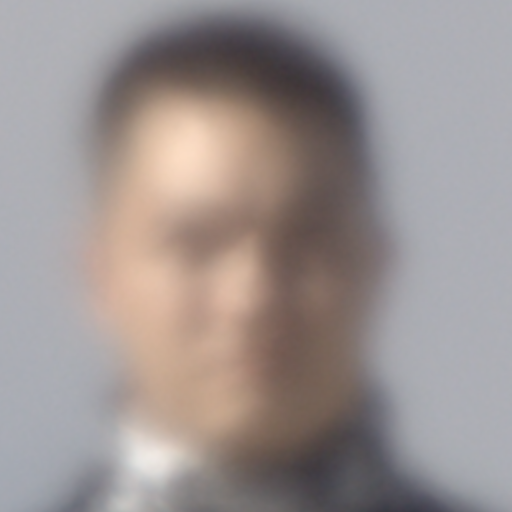}
    \tabularnewline
    \raisebox{0.3in}{\rotatebox[origin=t]{90}{\scriptsize \emph{AT-DDPM}}}&
    \includegraphics[width=0.22\linewidth]{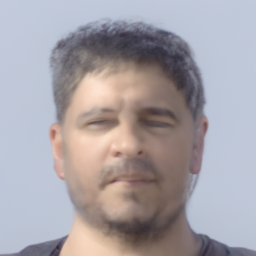} &
    \includegraphics[width=0.22\linewidth]{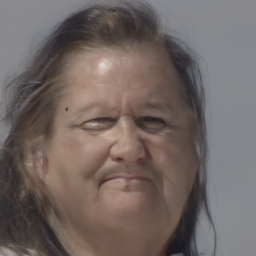} &
    \includegraphics[width=0.22\linewidth]{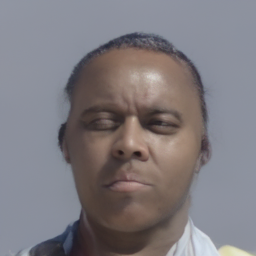} &
    \includegraphics[width=0.22\linewidth]{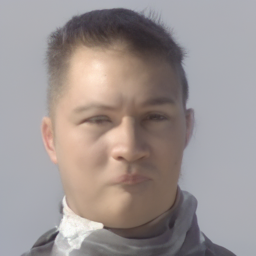}
    \tabularnewline
    \raisebox{0.3in}{\rotatebox[origin=t]{90}{\scriptsize \emph{Bi-noising (Ours)}}}&
    \includegraphics[width=0.22\linewidth]{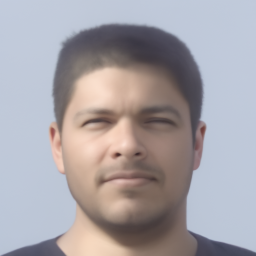} &
    \includegraphics[width=0.22\linewidth]{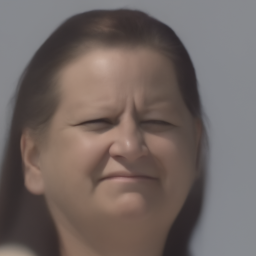} &
    \includegraphics[width=0.22\linewidth]{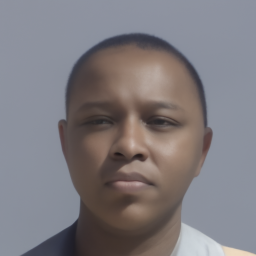} &
    \includegraphics[width=0.22\linewidth]{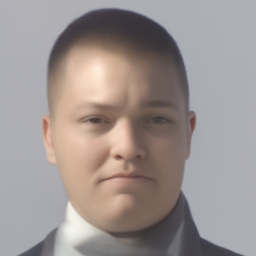}
    \tabularnewline
    \raisebox{0.3in}{\rotatebox[origin=t]{90}{\scriptsize \emph{Ground Truth}}}&
    \includegraphics[width=0.22\linewidth]{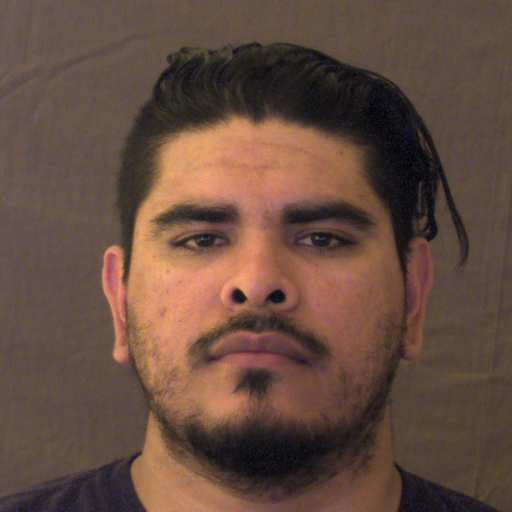} &
    \includegraphics[width=0.22\linewidth]{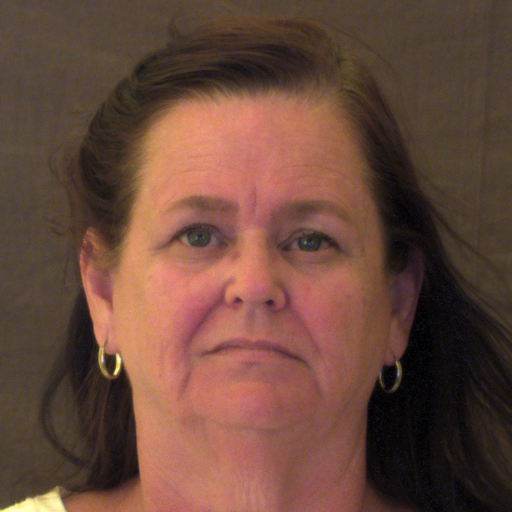} &
    \includegraphics[width=0.22\linewidth]{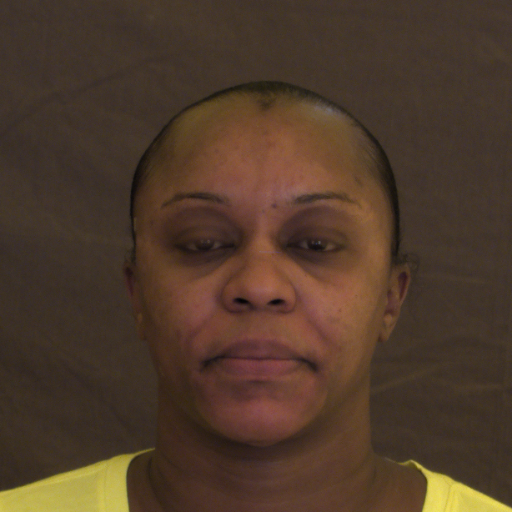} &
    \includegraphics[width=0.22\linewidth]{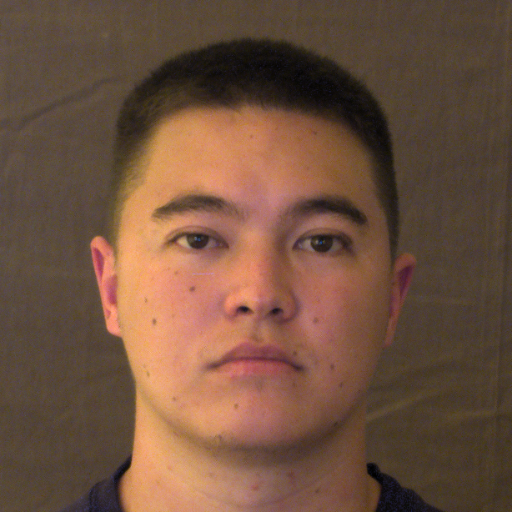}    
    \end{tabular}
    }
    }
    \vspace{-.5\baselineskip}
   \caption{\textbf{Atmospheric turbulence mitigation results corresponding to the LRFID dataset \cite{nair2022ddpm}.}}
    \label{fig:turb}
\vspace{-1.5\baselineskip}
\end{figure}

We plug the proposed bi-noising approach into the recent diffusion restoration work~\cite{nair2022ddpm} to demonstrate the applicability of our method on an extremely ill-posed atmospheric turbulence mitigation problem.  
Compared with the diffusion network with single noise conditioning, the results shown in Fig.~\ref{fig:turb} validate that our bi-noising method is able to remove the unnatural textures from the face images resulting from the incorrect denoising results.

\subsection{Design Analysis}
\noindent\textbf{Nonparametric v.s. Parametric Priors.}
Inspired by Ho et al.~\cite{ho2022classifier}, here we analysis the effect of alleviating complexity by parameterizing unconditional models into the conditional restoration model, denoted as \emph{Nonparametric Prior} in Tab.~\ref{tab:param}.
The compared methods use the same diffusion model but different guidance settings as the prior for fair comparisons. We showcase their performance and efficiency difference in the colorization task that was conducted using a single NVIDIA A6000 GPU.
Specifically, we use a single diffusion model that takes conditions for restoration, and it takes a null token $\varnothing$ for unconditional generation.
From Tab.~\ref{tab:param}, we can conclude that the nonparametric prior, \ie, w/o parametric, significantly reduces half of parameters for diffusion sampling, while the model suffers 0.26 dB performance drop compared with the parametric prior, \ie, Bi-Noising, that is our final setting.
The reason is that the null token increases the diffusion model training difficulty and thus the model fits worse than the unconditional model used in our final setting.
Compared with other concurrent works that utilize classifier guidance~\cite{dhariwal2021diffusion} and clip guidance~\cite{nichol2021glide}, our method outperforms them significantly with a slight increase in the number of parameters and running time.  This clearly demonstrates the benefits of our parametric prior that can encapsulate the low-level information distribution for restoration.

Fig.~\ref{fig:fid} further demonstrates the efficiency of our method that achieves better FID score after 2.0s denoising process. 

\begin{figure}[tp]
    \centering
    \setlength{\tabcolsep}{1pt}
    {\small
    \renewcommand{\arraystretch}{0.5} 
    \begin{tabular}{c c c c c c c c c}
    \captionsetup{type=figure, font=scriptsize}
    \raisebox{0.2in}{\rotatebox[origin=t]{90}{\scriptsize \emph{Input}}}&
    \includegraphics[width=0.25\linewidth]{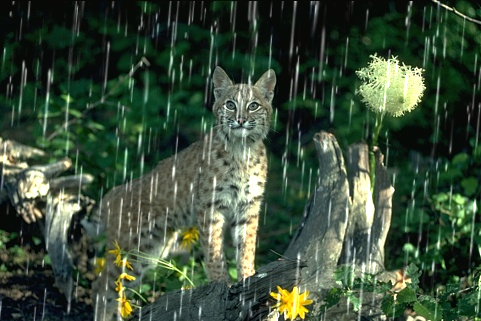}&
    \includegraphics[width=0.25\linewidth]{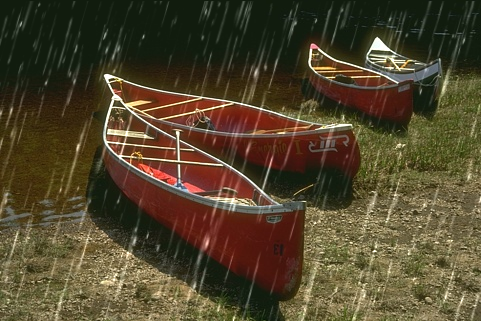}&
    \includegraphics[width=0.25\linewidth]{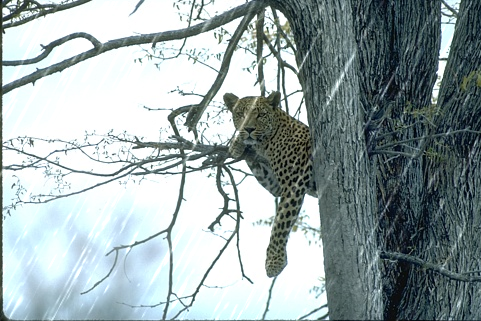}&
    \includegraphics[width=0.25\linewidth]{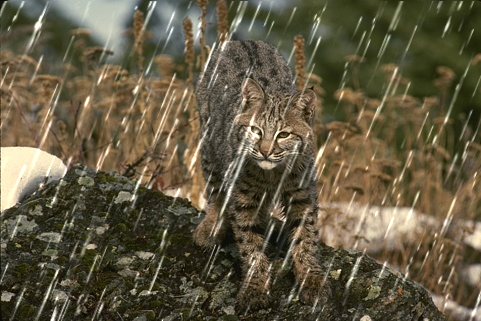}&

    \tabularnewline
    \raisebox{0.2in}{\rotatebox[origin=t]{90}{\scriptsize \emph{Ours}}}&
    \includegraphics[width=0.25\linewidth]{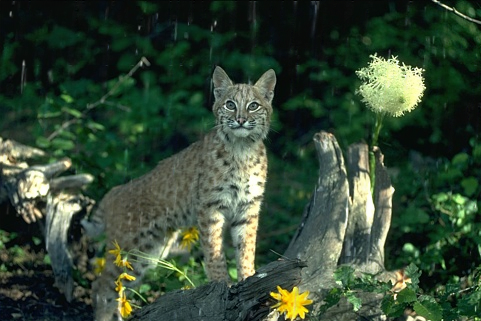}&
    \includegraphics[width=0.25\linewidth]{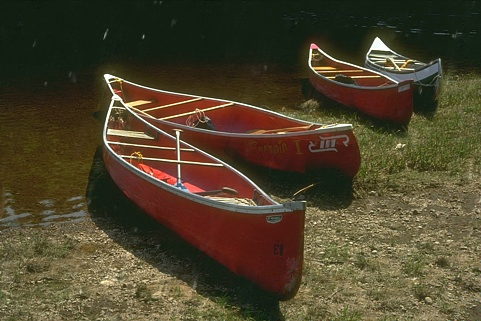}&
    \includegraphics[width=0.25\linewidth]{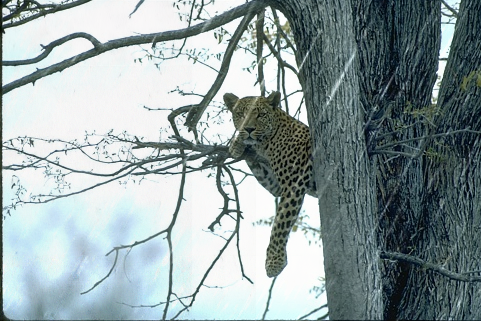}&
    \includegraphics[width=0.25\linewidth]{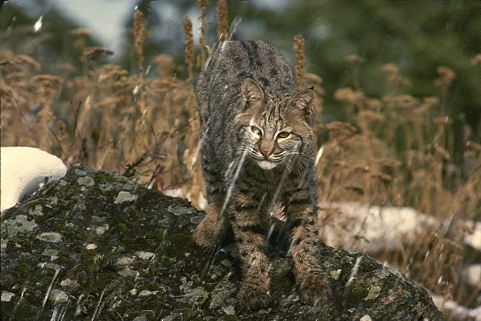}&

    \tabularnewline
    \raisebox{0.2in}{\rotatebox[origin=t]{90}{\scriptsize \emph{Ours + $\mathrm{L}_{\mathrm{corr}}$}}}&
    \includegraphics[width=0.25\linewidth]{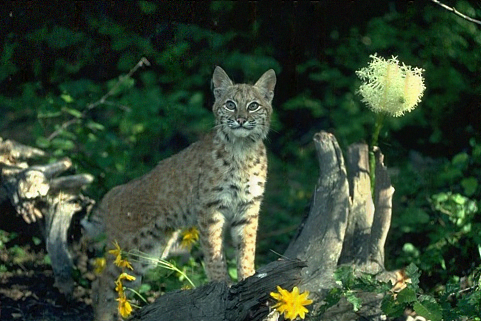}&
    \includegraphics[width=0.25\linewidth]{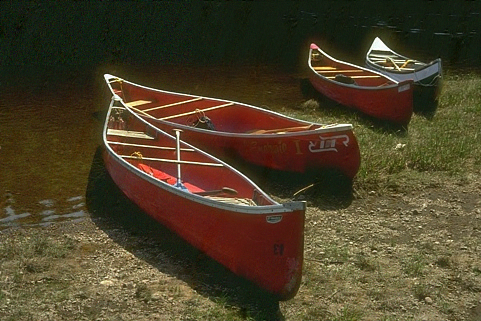}&
    \includegraphics[width=0.25\linewidth]{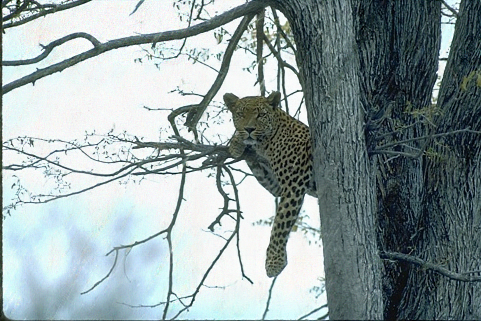}&
    \includegraphics[width=0.25\linewidth]{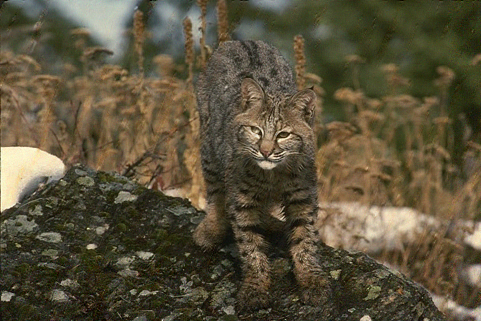}&
    \tabularnewline
    \raisebox{0.18in}{\rotatebox[origin=t]{90}{\scriptsize \emph{Ground Truth}}}&
    \includegraphics[width=0.25\linewidth]{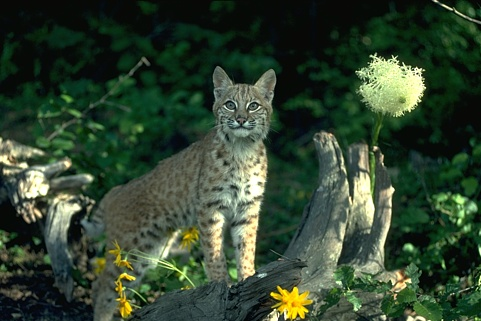}&
    \includegraphics[width=0.25\linewidth]{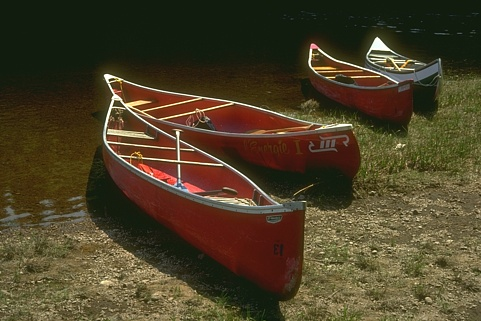}&
    \includegraphics[width=0.25\linewidth]{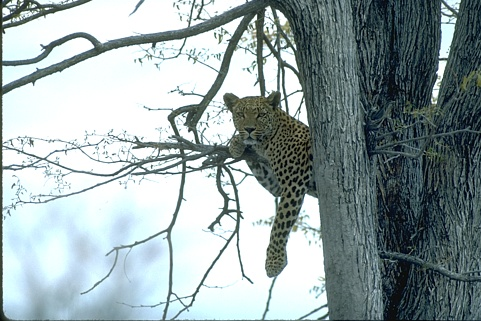}&
    \includegraphics[width=0.25\linewidth]{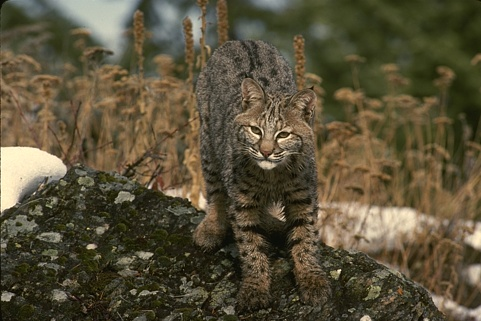}&

\end{tabular}}
\vspace{-.5\baselineskip}
\caption{\textbf{Deraining visual result comparisons that demonstrate the improvement brought by our $\mathrm{L}_{\mathrm{corr}}$ component.}}
\label{fig:rain800_supp}
\vspace{-1.5\baselineskip}
\end{figure}

\noindent\textbf{Priors Correlation.}
For the complex applications like deraining, we introduce additional correlation priors to further boost the final results.
In Tab.~\ref{tab:abj200}, we present the ablation study of the introduced correlation priors to demonstrate its effectiveness.
Since the rain streaks in the JORDER 200L dataset are relatively small, we choose the Rain800 dataset for a fair experiment.
The ablation starts with the base model and then adds the two priors one by one to show the improvements.
From the improved results due to $\mathrm{L}_{\mathrm{corr}}$, we can conclude that the introduced correlation priors allow our diffusion priors to better fit the probabilistic distribution of complex images, which ultimately benefits conditional generation with more realistic results.
The comparisons presented in Fig.~\ref{fig:rain800_supp} also visually validate the conclusion. 
% More results can be found in the supplementary file.

% \section{Discussion}

% \subsection{Efficiency Analysis}
% Efficiency has been a crucial issue of diffusion models since the models require multiple sampling timesteps until reaching the final results.
% Here we report the complexity comparison at the inference stage between different sampling methods and our DDPP in Table~\ref{tab:ef}.
% Even though our method doubles the running time with the same sampling timesteps, our efficiency superiority is revealed by using fewer sampling timesteps.
% Specifically, our method achieves 25dB performance in 3.40s, while the other compared method can only achieve less than 20dB performance with a similar running time of 3.48s. Therefore, the results suggest that our method can still achieve better efficiency by using fewer sampling timesteps without a large performance drop.

\begin{table}[]
  \centering
  \resizebox{1.0\linewidth}{!}{
  \begin{tabular}{lcccc}
    \toprule
    Method &  PSNR $\uparrow$  &SSIM $\uparrow$ & LPIPS $\downarrow$ & NIQE $\downarrow$  \\
    \midrule
    Rain Images (Input) & $26.70$ &$0.8439$&$0.2411$ &$4.131$ \\
    SR3 Diffusion~\cite{saharia2021image} &$31.45$ & $0.9091$ & $0.1779$&$3.588$   \\
     Ours & \emph{\underline{33.23}} & \emph{\underline{0.9505}} & \emph{\underline{0.1043}} & \emph{\underline{3.285}}  \\
     Ours + $\mathrm{L}_{\mathrm{corr}}$ & \textbf{33.90} & \textbf{0.9555} & \textbf{0.0972} & \textbf{3.232}  \\
    \bottomrule
    \end{tabular}}
    \vspace{-.1in}
    \caption{\textbf{Ablation study on the improvements brought by each introduced component.}}
    \label{tab:abj200}
    \vspace{-.5\baselineskip}
\end{table}

\section{Conclusion}
We explored ways in which one can utilize denoising diffusion probability model priors for improving image enhancement and restoration tasks.
The proposed way of integrating the stochastic priors into the deterministic conditioning denoising diffusion restoration model showed its superiority in colorization, face super-resolution, natural image super-resolution, and deraining tasks.
Compared with similar denoising diffusion-based restoration methods, the restored results of our introduced method achieved better color consistency and contain more fine-grained details.

\section{Acknowledgement}
This research is based upon work supported in part by the Ofﬁce of the Director of National Intelligence (ODNI), Intelligence Advanced Research Projects Activity (IARPA), via [2022-21102100005]. The views and conclusions contained herein are those of the authors and should not be interpreted as necessarily representing the ofﬁcial policies, either expressed or implied, of ODNI, IARPA, or the U.S. Government. The US. Government is authorized to reproduce and distribute reprints for governmental purposes notwithstanding any copyright annotation therein.

%%%%%%%%% REFERENCES
{\small
\bibliographystyle{ieee_fullname}
\bibliography{egbib}

\begin{thebibliography}{10}\itemsep=-1pt

\bibitem{abdal2019image2stylegan}
Rameen Abdal, Yipeng Qin, and Peter Wonka.
\newblock Image2stylegan: How to embed images into the stylegan latent space?
\newblock In {\em ICCV}, 2019.

\bibitem{abdal2020image2stylegan++}
Rameen Abdal, Yipeng Qin, and Peter Wonka.
\newblock Image2stylegan++: How to edit the embedded images?
\newblock In {\em CVPR}, 2020.

\bibitem{chan2021glean}
Kelvin~CK Chan, Xintao Wang, Xiangyu Xu, Jinwei Gu, and Chen~Change Loy.
\newblock Glean: Generative latent bank for large-factor image
  super-resolution.
\newblock In {\em CVPR}, 2021.

\bibitem{choi2021ilvr}
Jooyoung Choi, Sungwon Kim, Yonghyun Jeong, Youngjune Gwon, and Sungroh Yoon.
\newblock Ilvr: Conditioning method for denoising diffusion probabilistic
  models.
\newblock In {\em ICCV}, 2021.

\bibitem{creswell2018inverting}
Antonia Creswell and Anil~Anthony Bharath.
\newblock Inverting the generator of a generative adversarial network.
\newblock {\em IEEE TNNLS}, 2018.

\bibitem{dhariwal2021diffusion}
Prafulla Dhariwal and Alexander Nichol.
\newblock Diffusion models beat gans on image synthesis.
\newblock In {\em NeurIPS}, 2021.

\bibitem{goodfellow2014generative}
Ian Goodfellow, Jean Pouget-Abadie, Mehdi Mirza, Bing Xu, David Warde-Farley,
  Sherjil Ozair, Aaron Courville, and Yoshua Bengio.
\newblock Generative adversarial nets.
\newblock In {\em NeurIPS}, 2014.

\bibitem{gu2020image}
Jinjin Gu, Yujun Shen, and Bolei Zhou.
\newblock Image processing using multi-code gan prior.
\newblock In {\em CVPR}, 2020.

\bibitem{ho2020denoising}
Jonathan Ho, Ajay Jain, and Pieter Abbeel.
\newblock Denoising diffusion probabilistic models.
\newblock In {\em NeurIPS}, 2020.

\bibitem{ho2022classifier}
Jonathan Ho and Tim Salimans.
\newblock Classifier-free diffusion guidance.
\newblock {\em arXiv preprint arXiv:2207.12598}, 2022.

\bibitem{karras2017progressive}
Tero Karras, Timo Aila, Samuli Laine, and Jaakko Lehtinen.
\newblock Progressive growing of gans for improved quality, stability, and
  variation.
\newblock {\em arXiv preprint arXiv:1710.10196}, 2017.

\bibitem{karras2019style}
Tero Karras, Samuli Laine, and Timo Aila.
\newblock A style-based generator architecture for generative adversarial
  networks.
\newblock In {\em CVPR}, 2019.

\bibitem{karras2020analyzing}
Tero Karras, Samuli Laine, Miika Aittala, Janne Hellsten, Jaakko Lehtinen, and
  Timo Aila.
\newblock Analyzing and improving the image quality of stylegan.
\newblock In {\em CVPR}, 2020.

\bibitem{kingma2013auto}
Diederik~P Kingma and Max Welling.
\newblock Auto-encoding variational bayes.
\newblock {\em arXiv preprint arXiv:1312.6114}, 2013.

\bibitem{larochelle2011neural}
Hugo Larochelle and Iain Murray.
\newblock The neural autoregressive distribution estimator.
\newblock In {\em Proceedings of the fourteenth international conference on
  artificial intelligence and statistics}, 2011.

\bibitem{lau2020atfacegan}
Chun~Pong Lau, Hossein Souri, and Rama Chellappa.
\newblock Atfacegan: Single face image restoration and recognition from
  atmospheric turbulence.
\newblock In {\em 2020 15th IEEE International Conference on Automatic Face and
  Gesture Recognition (FG 2020)}, pages 32--39. IEEE, 2020.

\bibitem{ledig2017photo}
Christian Ledig, Lucas Theis, Ferenc Husz{\'a}r, Jose Caballero, Andrew
  Cunningham, Alejandro Acosta, Andrew Aitken, Alykhan Tejani, Johannes Totz,
  Zehan Wang, et~al.
\newblock Photo-realistic single image super-resolution using a generative
  adversarial network.
\newblock In {\em CVPR}, 2017.

\bibitem{liu2015deep}
Ziwei Liu, Ping Luo, Xiaogang Wang, and Xiaoou Tang.
\newblock Deep learning face attributes in the wild.
\newblock In {\em ICCV}, 2015.

\bibitem{lugmayr2022repaint}
Andreas Lugmayr, Martin Danelljan, Andres Romero, Fisher Yu, Radu Timofte, and
  Luc Van~Gool.
\newblock Repaint: Inpainting using denoising diffusion probabilistic models.
\newblock {\em arXiv preprint arXiv:2201.09865}, 2022.

\bibitem{mairal2009non}
Julien Mairal, Francis Bach, Jean Ponce, Guillermo Sapiro, and Andrew
  Zisserman.
\newblock Non-local sparse models for image restoration.
\newblock In {\em ICCV}, 2009.

\bibitem{mairal2007sparse}
Julien Mairal, Michael Elad, and Guillermo Sapiro.
\newblock Sparse representation for color image restoration.
\newblock {\em IEEE TIP}, 2007.

\bibitem{meng2022sdedit}
Chenlin Meng, Yutong He, Yang Song, Jiaming Song, Jiajun Wu, Jun-Yan Zhu, and
  Stefano Ermon.
\newblock {SDE}dit: Guided image synthesis and editing with stochastic
  differential equations.
\newblock In {\em ICLR}, 2022.

\bibitem{menon2020pulse}
Sachit Menon, Alexandru Damian, Shijia Hu, Nikhil Ravi, and Cynthia Rudin.
\newblock Pulse: Self-supervised photo upsampling via latent space exploration
  of generative models.
\newblock In {\em CVPR}, 2020.

\bibitem{miller2019data}
Kevin~J Miller, Bradley Preece, Todd~W Du~Bosq, and Kevin~R Leonard.
\newblock A data-constrained algorithm for the emulation of long-range
  turbulence-degraded video.
\newblock In {\em Infrared Imaging Systems: Design, Analysis, Modeling, and
  Testing XXX}, volume 11001, pages 204--214. SPIE, 2019.

\bibitem{nair2022ddpm}
Nithin~Gopalakrishnan Nair, Kangfu Mei, and Vishal~M Patel.
\newblock At-ddpm: Restoring faces degraded by atmospheric turbulence using
  denoising diffusion probabilistic models.
\newblock In {\em WACV}, 2022.

\bibitem{nichol2021glide}
Alex Nichol, Prafulla Dhariwal, Aditya Ramesh, Pranav Shyam, Pamela Mishkin,
  Bob McGrew, Ilya Sutskever, and Mark Chen.
\newblock Glide: Towards photorealistic image generation and editing with
  text-guided diffusion models.
\newblock {\em arXiv preprint arXiv:2112.10741}, 2021.

\bibitem{nichol2021improved}
Alexander~Quinn Nichol and Prafulla Dhariwal.
\newblock Improved denoising diffusion probabilistic models.
\newblock In {\em ICML}, 2021.

\bibitem{ramesh2022hierarchical}
Aditya Ramesh, Prafulla Dhariwal, Alex Nichol, Casey Chu, and Mark Chen.
\newblock Hierarchical text-conditional image generation with clip latents.
\newblock {\em arXiv preprint arXiv:2204.06125}, 2022.

\bibitem{richardson2021encoding}
Elad Richardson, Yuval Alaluf, Or Patashnik, Yotam Nitzan, Yaniv Azar, Stav
  Shapiro, and Daniel Cohen-Or.
\newblock Encoding in style: a stylegan encoder for image-to-image translation.
\newblock In {\em CVPR}, 2021.

\bibitem{russakovsky2015imagenet}
Olga Russakovsky, Jia Deng, Hao Su, Jonathan Krause, Sanjeev Satheesh, Sean Ma,
  Zhiheng Huang, Andrej Karpathy, Aditya Khosla, Michael Bernstein, et~al.
\newblock Imagenet large scale visual recognition challenge.
\newblock {\em IJCV}, 2015.

\bibitem{saharia2021image}
Chitwan Saharia, Jonathan Ho, William Chan, Tim Salimans, David~J Fleet, and
  Mohammad Norouzi.
\newblock Image super-resolution via iterative refinement.
\newblock {\em IEEE TPAMI}, 2022.

\bibitem{sohl2015deep}
Jascha Sohl-Dickstein, Eric Weiss, Niru Maheswaranathan, and Surya Ganguli.
\newblock Deep unsupervised learning using nonequilibrium thermodynamics.
\newblock In {\em ICML}, 2015.

\bibitem{song2020denoising}
Jiaming Song, Chenlin Meng, and Stefano Ermon.
\newblock Denoising diffusion implicit models.
\newblock {\em arXiv preprint arXiv:2010.02502}, 2020.

\bibitem{song2019generative}
Yang Song and Stefano Ermon.
\newblock Generative modeling by estimating gradients of the data distribution.
\newblock {\em NeurIPS}, 2019.

\bibitem{song2020score}
Yang Song, Jascha Sohl-Dickstein, Diederik~P Kingma, Abhishek Kumar, Stefano
  Ermon, and Ben Poole.
\newblock Score-based generative modeling through stochastic differential
  equations.
\newblock {\em arXiv preprint arXiv:2011.13456}, 2020.

\bibitem{wang2022semantic}
Weilun Wang, Jianmin Bao, Wengang Zhou, Dongdong Chen, Dong Chen, Lu Yuan, and
  Houqiang Li.
\newblock Semantic image synthesis via diffusion models.
\newblock {\em arXiv preprint arXiv:2207.00050}, 2022.

\bibitem{wang2021towards}
Xintao Wang, Yu Li, Honglun Zhang, and Ying Shan.
\newblock Towards real-world blind face restoration with generative facial
  prior.
\newblock In {\em CVPR}, 2021.

\bibitem{whang2021deblurring}
Jay Whang, Mauricio Delbracio, Hossein Talebi, Chitwan Saharia, Alexandros~G
  Dimakis, and Peyman Milanfar.
\newblock Deblurring via stochastic refinement.
\newblock {\em arXiv preprint arXiv:2112.02475}, 2021.

\bibitem{yang2017deep}
Wenhan Yang, Robby~T Tan, Jiashi Feng, Jiaying Liu, Zongming Guo, and Shuicheng
  Yan.
\newblock Deep joint rain detection and removal from a single image.
\newblock In {\em CVPR}, 2017.

\bibitem{yasarla2021learning}
Rajeev Yasarla and Vishal~M Patel.
\newblock Learning to restore images degraded by atmospheric turbulence using
  uncertainty.
\newblock In {\em 2021 IEEE International Conference on Image Processing
  (ICIP)}, pages 1694--1698. IEEE, 2021.

\bibitem{zhang2019image}
He Zhang, Vishwanath Sindagi, and Vishal~M Patel.
\newblock Image de-raining using a conditional generative adversarial network.
\newblock {\em IEEE TCSVT}, 2019.

\bibitem{zhu2016generative}
Jun-Yan Zhu, Philipp Kr{\"a}henb{\"u}hl, Eli Shechtman, and Alexei~A Efros.
\newblock Generative visual manipulation on the natural image manifold.
\newblock In {\em ECCV}, 2016.

\end{thebibliography}
}

\newpage
\appendix

\onecolumn
\section{Demo}
In order to provide a straightforward overview of our method, we have provided an online colorization demo and compared our bi-denoising process with the \emph{naive-diffusion} by visualizing their intermediate predicted $\bx_0$.
The online demo is accessible at \href{http://bi-noising.demohub.cc}{http://bi-noising.demohub.cc}.

\begin{figure}[htbp]
    \centering
    \includegraphics[width=.72\linewidth]{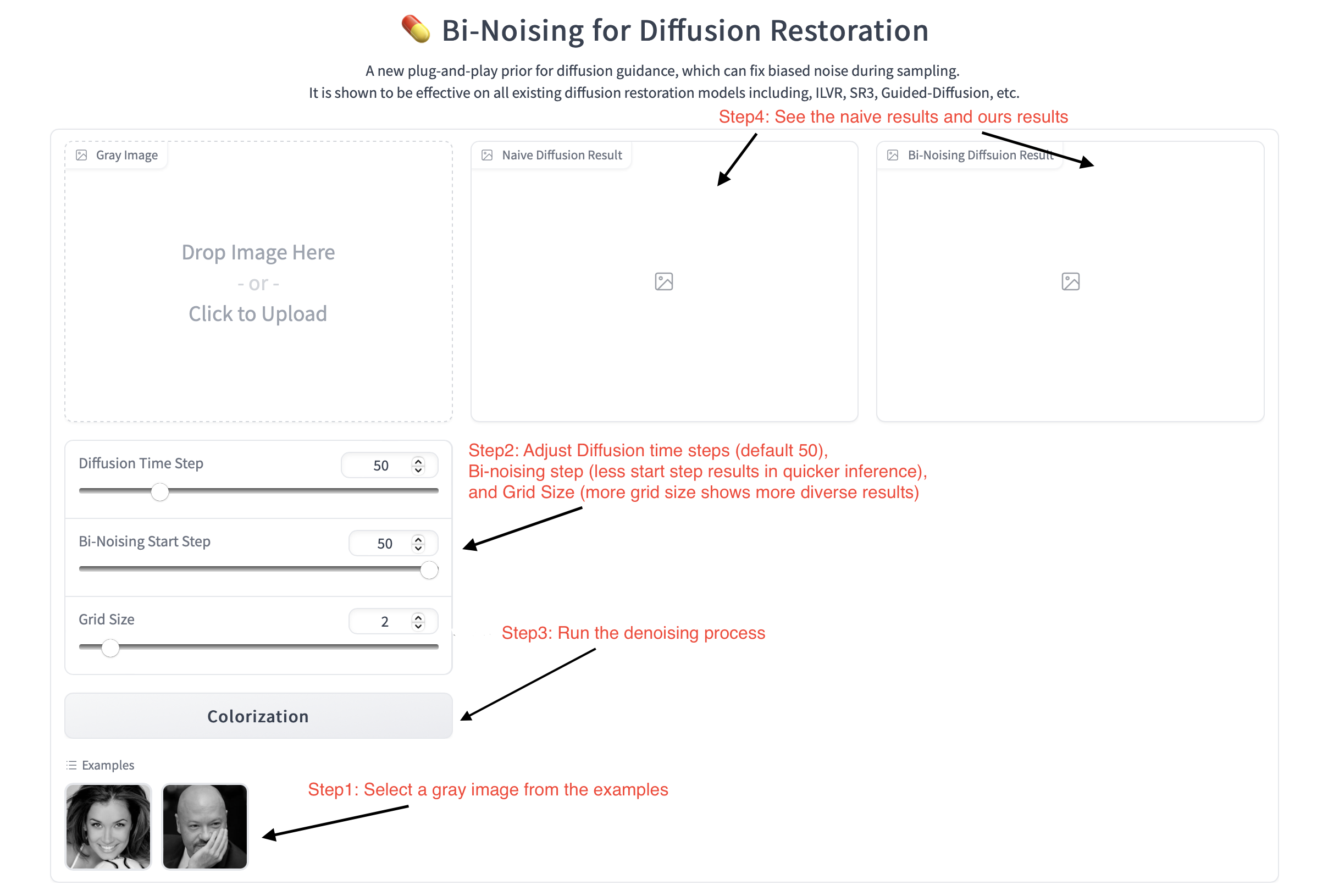}
    \caption{Screenshot of our online demo.}
\end{figure}

\begin{figure}[htbp]
    \centering
    \includegraphics[width=.72\linewidth]{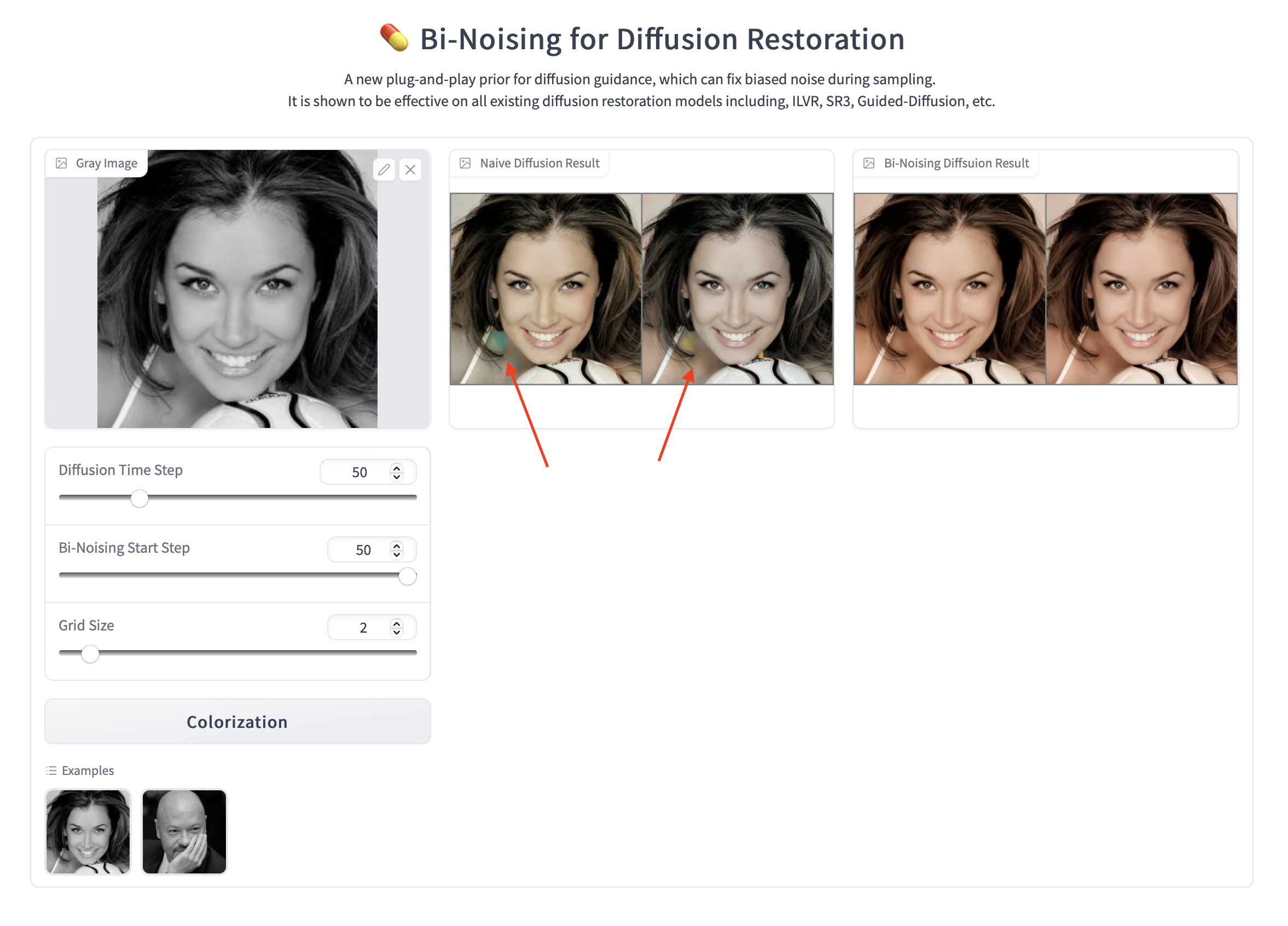}
    \caption{Screenshot of our online demo that shows artifacts in naive diffusion results, highlighted with red arrow.}
\end{figure}

\section{Manifold Correction using Diffusion Priors}
An alternate advantage of the proposed bi-noising diffusion is the effect of manifold correction. Consider any general restoration model$f(.))$ which maps from a domain $\mathcal{B}$ of the degraded image to a domain $\mathcal{A}$ of all natural images. The desired mapping function of this model  for any restoration task is to learn the mapping
\begin{align}
   b \in \mathcal{B}\xRightarrow[]{h(.)} a \in\mathcal{A}
\end{align}

\noindent But in real scenarios, because the restoration task is ill-posed, rather than learning the natural image manifold, a deep network learns an inverse function that merely removes the degradation effect. Let this manifold be denoted by $\mathcal{C}$. The mapping function hence learned is
\begin{align}
   b \in  \mathcal{B}\xRightarrow[]{f(.)} a \in\mathcal{C}
\end{align}

For example, for a restoration model trained for the task of face super-resolution, for an input (b), the network could create an output (c)  that is the image of a disoriented face rather than an image in the manifold of faces. Theoretically, if we utilize any generative model for the restoration, the model should be able to achieve the ideal mapping. But often, the model learns the more complex problem of removing the degradation than learning to map to the domain of natural images. This is because it is difficult to reach the solution corresponding to the global optimum. In any generic restoration method, this deviation from the natural manifold can be corrected by adding a correction network that learns the mapping from domain $\mathcal{C}$ to $\mathcal{A}$. Unlike all other models, diffusion models contain a flexible model structure where intermediate latent variables can be accessed. This enables a manifold correction during inference time alone with explictly training a network to map from the generated manifold $\mathcal{C}$ to the natural manifold $\mathcal{A}$. Hence in our work, we exploit this property and perform the manifold correction to the domain of natural images through an additional step that utilizing an  unconditional model. Consider a CDP trained for any restoration task denoted by $f_{\phi}(c_t,b,t)$. During inference, the restored sample $c_T$ is generated through the cascade of steps
\begin{align}
    f_{\phi}(c_0,b,t)\rightarrow  f_{\phi}(c_1,b,t),.... \rightarrow f_{\phi}(c_T,b,t)
\end{align}
or equivalently,
\begin{align}
    c_0 \xrightarrow[]{f_{\phi}(.)} c_1 ,... \xrightarrow[]{f_{\phi}(.)} c_T 
\end{align}

Here, ${c_1,...,c_T}$ denotes the intermediate diffusion outputs of an image $c_T$in the manifold $\mathcal{C}$ that can be reconstructed from a degraded b. As mentioned before, the function $f$ will not always map to the domain $\mathcal{A}$ of natural images. Hence we add an unconditional model $g_{\theta}(.)$ that does the task of aligning the manifold of the generated image to the manifold of natural images. The sequence of operations is as follows
\begin{align}
    f_{\phi}(c_0,b,t)\rightarrow  g_{\theta}(c_1,t),.... f_{\phi}(a_{T-1},b,t)\rightarrow g_{\theta}(c_T,t) \\
    c_0 \xrightarrow[]{f_{\phi}(.)} c_1\xrightarrow[]{g_{\theta}(.)} a_1\xrightarrow[]{f_{\phi}(.)}c_2,... \xrightarrow[]{f_{\phi}(.)} c_T \xrightarrow[]{g_{\theta}(.)} a_T
\end{align}
 ${a_1,...,a_T}$ denotes the intermediate diffusion outputs of an image $a_T$ in  manifold $\mathcal{A}$.

\section{Turbulence Removal}
Here we provide the quantitative evaluation of our method on the turbulence removal benchmark LRFID dataset.
Compared with the other methods, ours not only has achieved better performance in the sample quality in terms of LPIPS but also better fidelity in terms of face recognition accuracy Top-1 and Top-3.

\begin{table}[htbp!]
% \tiny
% \vspace{-0.5cm}
	\begin{center}
% 			\vspace{-10pt}
		% \resizebox{\textwidth}{!}{
			\begin{tabular}{|c | c c c|}
			 \hline
			  \multirow{1}{*}{ Dataset }&   \multicolumn{3}{c|}{LRFID dataset\cite{miller2019data}}\\
				\hline
				Metric&LPIPS$(\downarrow)$&Top-1$(\uparrow)$&Top-3$(\uparrow)$\\		\hline
                % \hline
				% GT&0&0&100.0&100.0&100.0\\
				degraded&0.6293&35.3&62.2\\
				\hline
				    &\multicolumn{3}{c|}{CNN based models}\\
				    %   \hline
				MPRNET\cite{yasarla2021learning}& 0.5755&34.1&64.6\\	
				ATNet\cite{yasarla2021learning}& 0.6128&36.5&64.6\\
								\hline
				       &\multicolumn{3}{c|}{GAN based models}\\
				    %   \hline
		ATFaceGAN\cite{lau2020atfacegan}& 0.6300&\underline{\emph{47.5}}&\underline{\emph{65.8}}\\
				\hline
				      & \multicolumn{3}{c|}{Diffusion models}\\	
				ILVR Diffusion \cite{choi2021ilvr}&\emph{\underline{0.5661}}&31.7&59.7\\
				Bi-Noising Diffusion (Ours) &\textbf{0.5500}&\textbf{48.7}&\textbf{73.1}\\
			
				% FID&&&&&&&\\
				% LPIPS&&&&&&&\\
				\hline
			\end{tabular}
		% }
					\caption{Quantitative results on on real world turbulence degraded datasets: LRFID dataset \cite{miller2019data}
			\label{table:quant}}
	\end{center}
	\vspace{-0.8cm}
\end{table}

\end{document}